%% file: acl_latex.tex
\pdfoutput=1
\documentclass[11pt]{article}

\usepackage[final]{acl}

\usepackage{times}
\usepackage{latexsym}
\usepackage{microtype}
\usepackage{graphicx}
\usepackage{subcaption}
\usepackage{booktabs}
\usepackage{amsmath,amssymb}
\usepackage{bbm}
\usepackage{etoolbox,lipsum}
\usepackage{multirow}
\usepackage[ruled,vlined]{algorithm2e}
\usepackage[T1]{fontenc}
\usepackage[utf8]{inputenc}

\usepackage{microtype}

\usepackage{inconsolata}

\title{From Zero to Hero: Cold-Start Anomaly Detection}

\author{\vspace{2mm} \hspace{2mm} Tal Reiss$^{1}$
\hspace{2mm} 
George Kour$^{2}$
\hspace{2mm}
Naama Zwerdling$^{2}$ 
\hspace{2mm}
Ateret Anaby-Tavor$^{2}$
\hspace{2mm}
Yedid Hoshen$^{1}$ 
\\
\vspace{2mm}
\normalsize{$^{1}$The Hebrew University of Jerusalem}
\hspace{7mm} \normalsize{$^{2}$IBM}
\\
\normalsize{\url{https://github.com/talreiss/ColdFusion}}
}

\begin{document}
\maketitle

\input{sec/0_abstract}    
\input{sec/1_intro}
\input{sec/2_cold_start}
\input{sec/3_method}
\input{sec/4_experiments}

\input{sec/5_conclusion}
\input{sec/6_limitations}
\input{sec/7_ethics}
\input{sec/8_ack}

\bibliography{custom}

\clearpage
\appendix
\input{sec/X_appendix}

\end{document}

%% file: sec/0_abstract.tex
\begin{abstract}
When first deploying an anomaly detection system, e.g., to detect out-of-scope queries in chatbots, there are no observed data, making data-driven approaches ineffective. Zero-shot anomaly detection methods offer a solution to such "cold-start" cases, but unfortunately they are often not accurate enough. This paper studies the realistic but underexplored \textit{cold-start} setting where an anomaly detection model is initialized using zero-shot guidance, but subsequently receives a small number of contaminated observations (namely, that may include anomalies). The goal is to make efficient use of both the zero-shot guidance and the observations. We propose ColdFusion, a method that effectively adapts the zero-shot anomaly detector to contaminated observations. To support future development of this new setting, we propose an evaluation suite consisting of evaluation protocols and metrics.
\end{abstract}

%% file: sec/1_intro.tex
\section{Introduction}
Anomaly detection methods aim to flag data that violate accepted norms. For example, a customer support chatbot may be designed to answer queries about particular intents (in-scope) but not about other intents (out-of-scope). Unlike related tasks such as out-of-scope intent discovery and classification, which rely on large labeled in-scope data, anomaly detection approaches relax the labeling assumption and treat the problem as a one-class classification task \citep{lin2020discovering,zhang2021discovering,mou2022disentangled,zheng2020out,zhan2021out,lin2019deep,zeng2021modeling,zhang2021deep,xu2020deep}. Most anomaly detection methods \citep{panda,time_series,pretrained2_cvpr23} require previous observations for training and are effective when many past observations are available. Such methods are not effective for systems just after deployment, as they lack access to any past observations. Zero-shot anomaly detection \citep{zero_winclip,zero_MuSc,zero_anomalyclip} uses descriptions of the normal classes and does not require training data. While zero-shot methods can be used for freshly deployed systems, they result in reduced accuracy as the descriptions often fail to properly express the distribution of real data.

We explore the \textit{cold-start} setting which provides two types of guidance: i) a textual description of each normal class, serving as initial guidance, such as predefined topic names in chatbot systems; ii) a stream of $t$ contaminated observations (that may include anomalies), e.g., real user queries. It is particularly relevant in real-world applications where, shortly after deployment, a short stream of user queries becomes available but the queries are not labeled into intent types and some of them are out-of-scope. To our knowledge, the only work that deals with a similar setting \citep{zero_winclip} assumes prior knowledge of anomalies, that observations come from a single normal class and that they are not contaminated by anomalies.

To tackle the cold-start setting, we present ColdFusion, a method for adapting a zero-shot model given the distribution of a limited observation stream. Our method is very effective, achieving considerably better results than pure zero-shot and observation-based methods. To encourage future research on this promising new setting, we provide evaluation protocols and metrics. \hfill \hfill \break
Our contributions are:
\begin{enumerate}
    \vspace{-0.25em}
    \item Proposing the new setting of \textit{cold-start anomaly detection}.
    \vspace{-0.25em}
    \item Presenting \textit{ColdFusion} for tackling the setting.
    \vspace{-1.55em}
    \item Introducing a dedicated evaluation suite consisting of evaluation protocols and metrics.
\end{enumerate}

%% file: sec/2_cold_start.tex
\section{Cold-Start Anomaly Detection}
\textbf{Task Definition.} In the cold-start setting, a model has access to $K$ class descriptions $\mathcal{D}_{\text{prior}}=\{c_1,c_2,...,c_K\}$ and a stream of $t$ observations $\mathcal{D}_{t} = \{x_1,x_2,...,x_t\}$, where $t$ is small. We denote the percentage of anomalous observations as the \textit{contamination ratio} $r\%$. An observation $x$ either comes from one of the $K$ normal classes or is anomalous, but we do \textit{not} have access to the class or anomaly label. The task is to learn a model $S$ to map each training sample $x$ to an anomaly score such that high values indicate anomalous samples. 

\textbf{Application to chatbots.} Our practical motivation is identifying out-of-scope queries in a recently deployed chatbot. We observe a stream of queries sent to the chatbot, as well as descriptions of all allowed intents. At time step $t+1$, we leverage both $\mathcal{D}_{t}$ and $\mathcal{D}_{\text{prior}}$ to classify a given query $x_{t+1}$ as in-scope (INS) or out-of-scope (OOS).

%% file: sec/3_method.tex
\begin{figure}[t]
  \begin{center}
    \begin{tabular}{c}
    \includegraphics[scale=1]{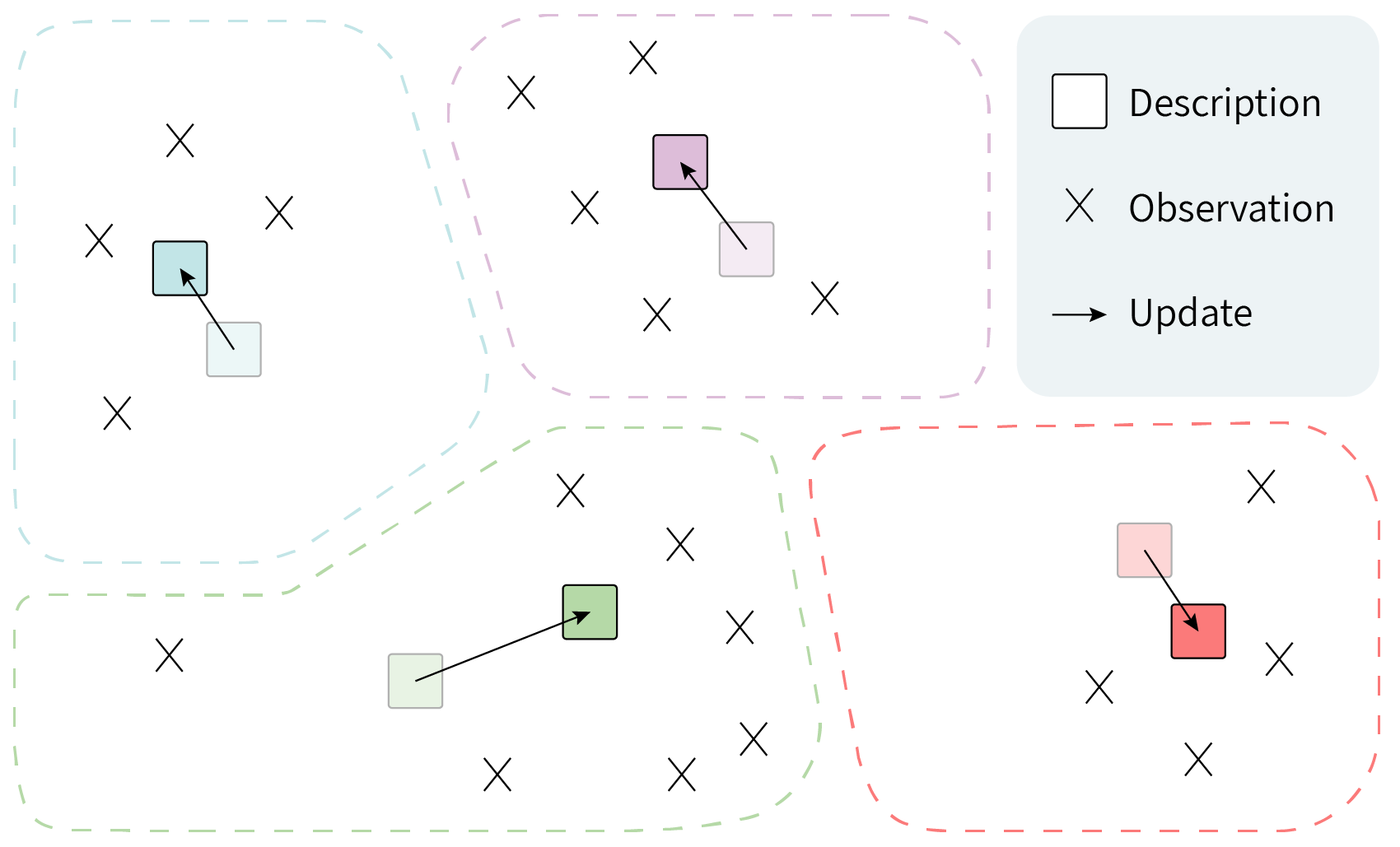} 
    \end{tabular}
  \end{center}
  \vspace{-0.75em}
    \caption{ColdFusion assigns each of the $t$ observations to their nearest class, then adapts the embeddings of each class towards the assigned observations.}
    \label{fig:coldfusion}
    \vspace{-0.5em}
\end{figure}

\begin{table*}[ht]
\centering
\resizebox{1.0\linewidth}{!}{%
\begin{tabular}{llcccccccccccc}
\toprule
\parbox[t]{4mm}{\multirow{1}{*}{\rotatebox[origin=c]{90}{\small{Encoder}}}}  & \multirow{3}{*}{Method} & \multicolumn{3}{c}{$\text{AUC}^2_{10\%}$} & \multicolumn{3}{c}{$\text{AUC}^2_{25\%}$} & \multicolumn{3}{c}{$\text{AUC}^2_{50\%}$}& \multicolumn{3}{c}{$\text{AUC}^2_{100\%}$}  \\
  \cmidrule(lr){3-5} \cmidrule(lr){6-8} \cmidrule(lr){9-11} \cmidrule(lr){12-14}
  & & B77 & C-Bank & C-Cards &  B77 & C-Bank & C-Cards & B77 & C-Bank & C-Cards & B77 & C-Bank & C-Cards \\
 \cmidrule(lr){1-1} \cmidrule(lr){2-2} \cmidrule(lr){3-3} \cmidrule(lr){4-4} \cmidrule(lr){5-5} \cmidrule(lr){6-6} \cmidrule(lr){7-7} \cmidrule(lr){8-8} \cmidrule(lr){9-9} \cmidrule(lr){10-10} \cmidrule(lr){11-11} \cmidrule(lr){12-12} \cmidrule(lr){13-13} \cmidrule(lr){14-14}
\parbox[t]{4mm}{\multirow{3}{*}{\rotatebox[origin=c]{90}{GTE}}} & ZS & 78.9 & \textbf{83.1} & 81.8 & 78.9 & 83.1 & 81.8 & 78.9 & 83.1 & 81.8 & 78.9 & 83.1 & 81.8  \\
& DN2 & 76.7 & 64.8 & 70.0 & 76.2 & 76.0 & 75.6 & 75.9 & 80.2 & 79.6 & 75.3 & 82.2 & 80.2  \\
& ColdFusion & \textbf{81.7} & 82.3 & \textbf{84.8} & \textbf{81.8} & \textbf{87.0} & \textbf{87.3} & \textbf{81.9} & \textbf{88.6} & \textbf{88.7} & \textbf{82.3} & \textbf{89.2} & \textbf{89.0}  \\
\cmidrule(lr){1-14}
\parbox[t]{4mm}{\multirow{3}{*}{\rotatebox[origin=c]{90}{MPNET}}} & ZS & 81.8 & 82.7 & 80.1 & 81.8 & 82.7 & 80.1 & 81.8 & 82.7 & 80.1 & 81.8 & 82.7 & 80.1 \\
& DN2 & 78.3 & 69.7 & 69.8 & 78.2 & 78.9 & 76.9 & 77.6 & 82.3 & 80.9 & 76.3 & 83.6 & 81.1  \\
& ColdFusion & \textbf{83.3} & \textbf{84.4} & \textbf{84.1} & \textbf{82.8} & \textbf{87.8} & \textbf{86.0} & \textbf{82.8} & \textbf{88.8} & \textbf{87.8} & \textbf{83.0} & \textbf{89.4} & \textbf{88.3}  \\
\bottomrule
\end{tabular}
}
\caption{$\text{AUC}^2_{\tilde{t}}$ results, with contamination of $r=5\%$. Best results are in bold.}
\label{tab:co}
\end{table*}

\section{Method}
\label{sec:method}

\subsection{Recap: Zero-Shot Anomaly Detection}
\label{sec:zs}
Zero-shot (ZS) anomaly detection maps each data point $x$ to an anomaly score $S(x)$. Notably, ZS methods do not require past data, instead they are guided by a set of distinct normal class names $\{c_1,c_2,...,c_K\}$ provided by the user. A pre-trained feature extractor $\phi$ maps each of the class names $c_k$, and observations $x_t$ to deep embeddings $\phi(c_k)$ and $\phi(x_t)$. It then computes the distance $d$ (often $L_2$ or Cosine) between the embeddings of the example and each of the class names. The final anomaly score is given by the distance to the nearest class:
\vspace{-0.25em}
\begin{equation}
S_{zs}(x) = \min_k \{d(\phi(x),\phi(c_k))\}_{k=1}^K
\vspace{-0.5em}
\end{equation}
High anomaly scores serve as indicators of anomalies. The anomaly score can be converted to a binary label by choosing a threshold $\alpha$ such that $y=0$ if $S(x) < \alpha$ and $y=1$ if $S(x) \geq \alpha$.

Zero-shot anomaly detection can be used for OOS query detection by first specifying a set of allowed intents. Then a deep encoder extracts the embeddings of the target user query and intent descriptions. Finally, the method labels the user query as OOS if it is far from all allowed intent names.

\begin{algorithm}[t]
\caption{ColdFusion}
\label{alg:coldfusion_algorithm}
\KwIn{$\mathcal{D}_{\text{prior}}$, $\mathcal{D}_{t}$, $p$, query $x$.}
\KwOut{Anomaly score $S_{adapt}(x)$.}
\textbf{Step 1:} Encode class descriptions and observations: $\phi(\mathcal{D}_{\text{prior}})$, $\phi(\mathcal{D}_{t})$\;
\textbf{Step 2:} Assign observations to classes based on nearest class descriptor: $a(x) = \arg\min_k \{d(\phi(x),\phi(c_k))\}_{k=1}^K$\;
\textbf{Step 3:} Adapt class embeddings: $z_k = \text{median}(\phi(c_k),\{\phi(x)|a(x)=k\})$\;
\textbf{Step 4:} Compute anomaly score for $x$: $S_{adapt}(x) = \min_k\{d(\phi(x),z_k)\}_{k=1}^K$\;
\end{algorithm}
\vspace{-0.5em}

\subsection{Limitations of Existing Methods}
In practice, it is impossible to provide perfect class descriptions, and therefore zero-shot anomaly detection often does not achieve sufficient accuracy. On the other hand, if the number of observations is limited, observation-based anomaly detection methods, such as $K$-nearest neighbors, struggle for three key reasons: i) the observations may not include all in-scope classes; ii) it is hard to estimate the true distribution of normal data from a few samples; iii) the observations may be contaminated by anomalies. Empirically, observation-based methods underperform ZS methods for small $t$ (see Tab.~\ref{tab:co} and Fig.~\ref{fig:over_time}).

\subsection{Our Method: ColdFusion}
\label{sec:coldfusion}
To bridge the gap between ZS and observation-based methods, we propose \textit{ColdFusion} (illustrated in Fig.~\ref{fig:coldfusion}), a method for cold-start anomaly detection using domain adaptation. It improves ZS anomaly detection using the $t$ observations in two key stages: i) assigning observations to classes; ii) adapting ZS class embeddings based on the assigned observations.

\textbf{Assignment.} We assign each of the $t$ observations  to the nearest class as measured in the feature space $\phi$. We denote the class assignment of observation $x$ as $a(x)$. More formally:
\vspace{-0.25em}
\begin{equation}
    a(x) = \arg\min_k \{d(\phi(x),\phi(c_k))\}_{k=1}^K
    \vspace{-0.25em}
\end{equation}
We further define $\mathcal{C}_k$, the set of all observations assigned to class $k$ as $\mathcal{C}_k = \{\phi(x)|a(x)=k\}$.

\textbf{Adaptation.} We now adapt each class embedding by considering both the initial class description and the assigned observations. Concretely, the adapted code for each class is the median of the set containing the embedding of the class descriptions and the embeddings of all assigned observations:
\begin{equation}
    z_k = median(\{\phi(c_k)\} \cup \mathcal{C}_k)
    % \vspace{-0.35em}
\end{equation}
We chose the median and not mean for contamination robustness. Note that this step will not modify the embedding of classes with no observations.

\textbf{Anomaly scoring.} ColdFusion uses the same anomaly scoring as ZS except that the class codes are the adapted $\{z_k\}_{k=1}^K$ instead of the encoding of the original description i.e., $S_{adapt}(x) = \min_k \{d(\phi(x_{t+1}),z_k)\}_{k=1}^K$.

%% file: sec/4_experiments.tex
\begin{figure*}[t]
     \centering
     \begin{subfigure}[b]{0.3275\textwidth}
         \centering
         \includegraphics[width=\textwidth]{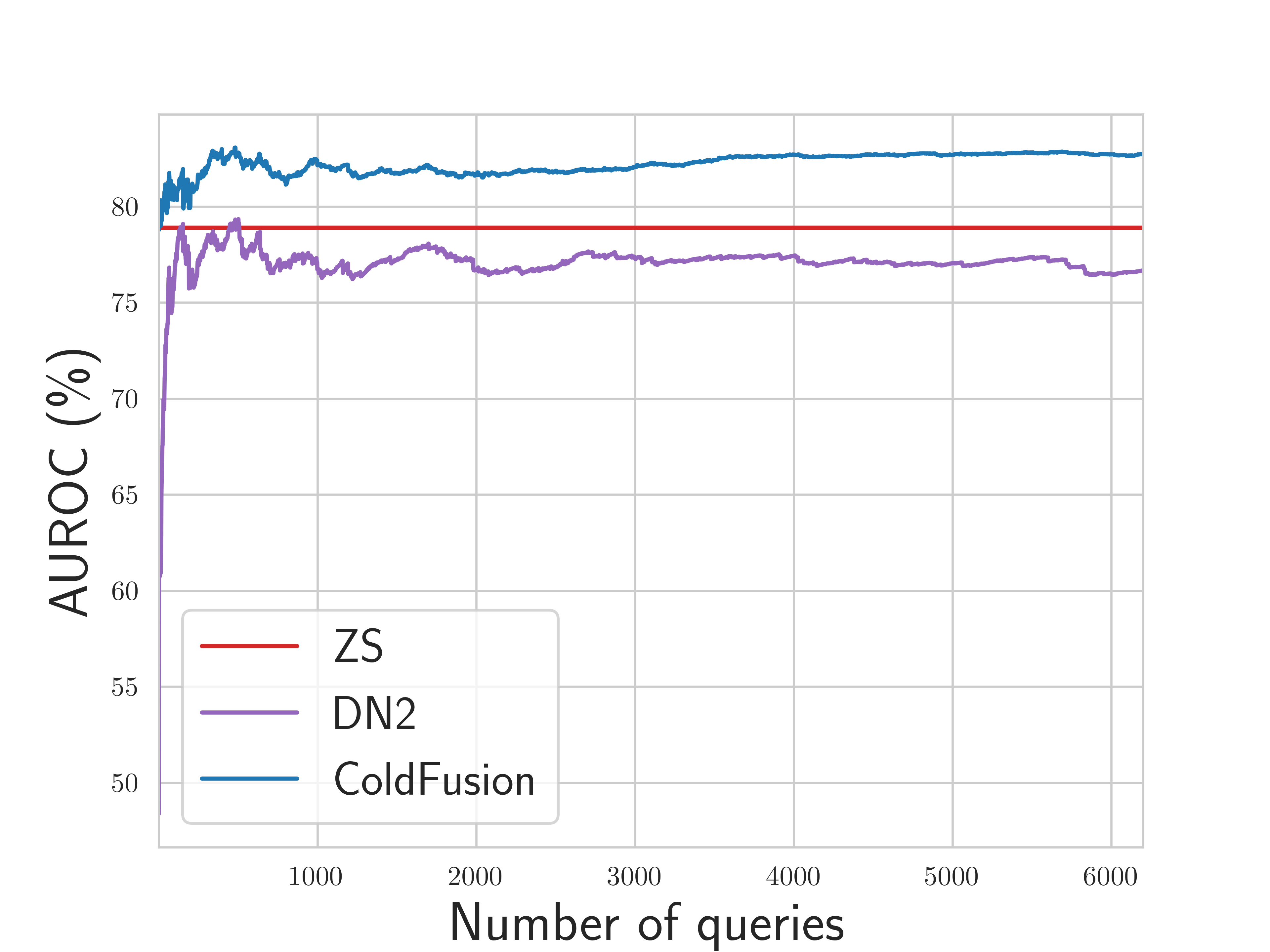}
         \vspace{-1em}
         \caption{Banking77-OOS}
     \end{subfigure}
     \hfill
     \begin{subfigure}[b]{0.3275\textwidth}
         \centering
         \includegraphics[width=\textwidth]{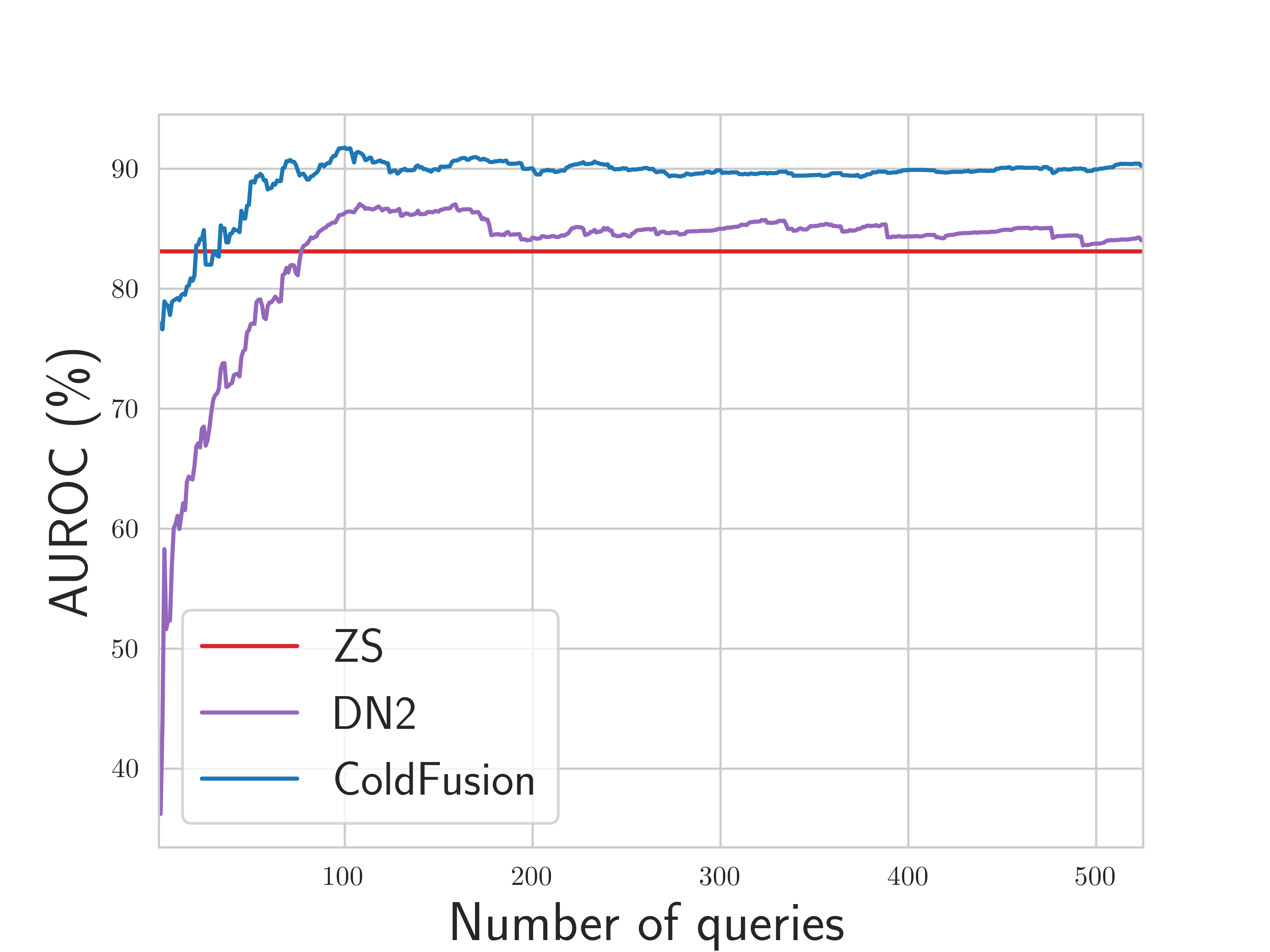}
         \vspace{-1em}
         \caption{CLINC-Banking}
     \end{subfigure}
     \hfill
     \begin{subfigure}[b]{0.3275\textwidth}
         \centering
         \includegraphics[width=\textwidth]{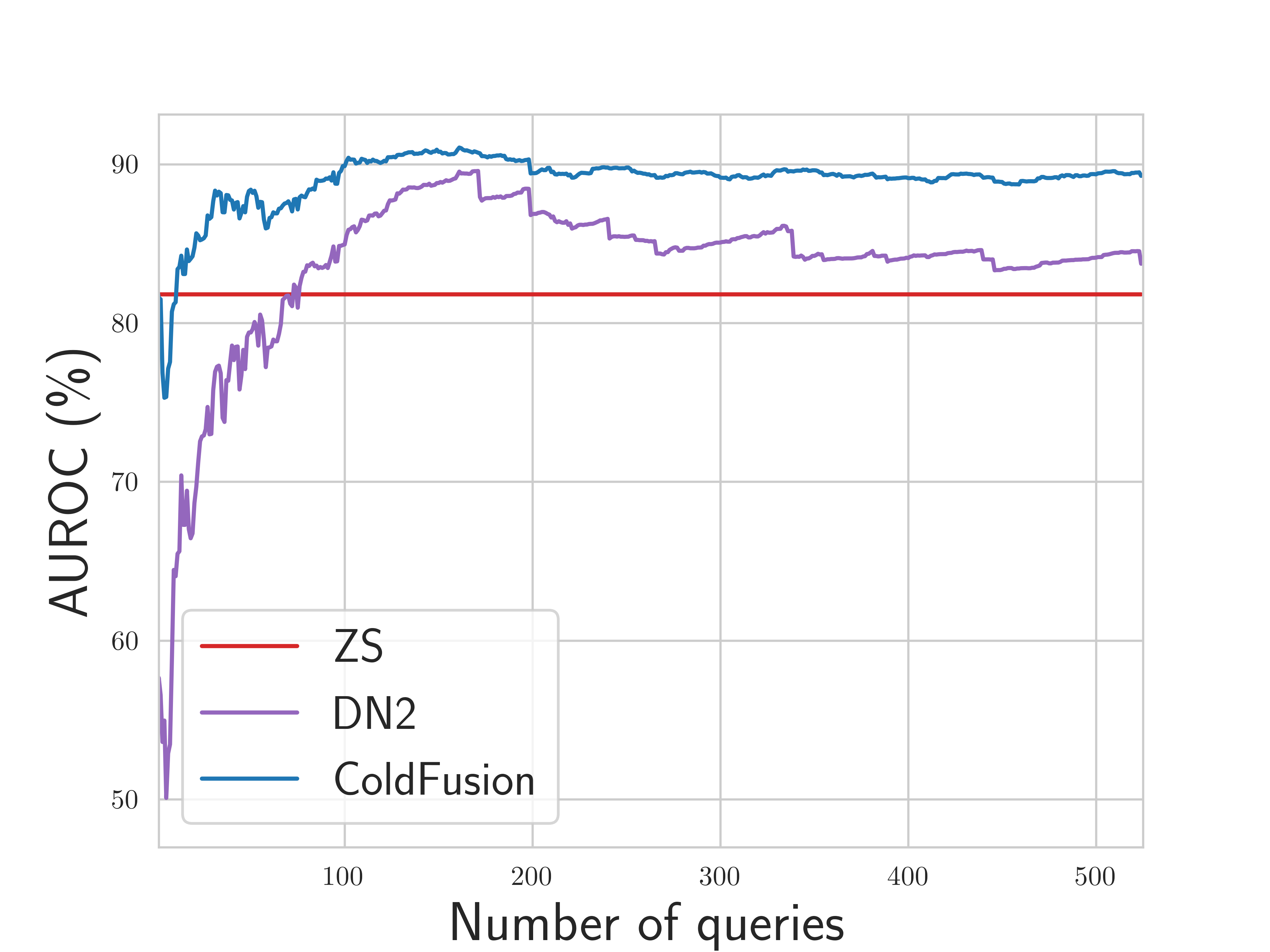}
         \vspace{-1em}
         \caption{CLINC-Credit\_Cards}
     \end{subfigure}
        \caption{Performance trends with contamination $r=5\%$ using the GTE model over time demonstrate the superiority of our ColdFusion method over other baseline approaches.}
        \label{fig:over_time}
\end{figure*}

\begin{table}[t]
\centering
\resizebox{1.0\linewidth}{!}{%
\begin{tabular}{lcccccc}
\toprule
  \multirow{3}{*}{Method} & \multicolumn{3}{c}{$\text{AUC}^2_{10\%}$} & \multicolumn{3}{c}{$\text{AUC}^2_{25\%}$}  \\
  \cmidrule(lr){2-4} \cmidrule(lr){5-7}
  & B77 & C-Bank & C-Cards &  B77 & C-Bank & C-Cards  \\
\cmidrule(lr){1-1} \cmidrule(lr){2-2} \cmidrule(lr){3-3} \cmidrule(lr){4-4} \cmidrule(lr){5-5} \cmidrule(lr){6-6} \cmidrule(lr){7-7} 
$K$-means  & 80.0 & 79.2 & 83.7 & 78.9 & 84.0 & 87.0     \\
Mean & 81.6 & 80.8 & 84.7 & 81.7 & 86.4 & \textbf{87.5} \\
MI & 81.6 & \textbf{82.3} & \textbf{84.8} & \textbf{81.8} & \textbf{87.0} & 87.1 \\
Median & \textbf{81.7} & \textbf{82.3} & \textbf{84.8} & \textbf{81.8} & \textbf{87.0} & 87.3 \\
\bottomrule
\end{tabular}
}
\caption{$\text{AUC}^2_{\tilde{t}}$ results, with contamination of $r=5\%$ using the GTE model. MI refers to multiple iterations with median adaptation. Best results are in bold.}
\label{tab:adaptation}
\end{table}

\section{Experiments}

\textbf{Experimental setting.} Our experiments simulate the deployment of an OOS query detection system. We first randomly sort the queries so that each query has a unique time $t$, modeling a query stream. At each time $t$, we train a model using the $t$ available observations and the $K$ class names, and evaluate the model on the entire test set.

\noindent\textbf{Datasets.} We use three evaluation datasets, Banking77-OOS and CLINC-OOS segmented into CLINC-Banking and CLINC-Credit\_Cards. Banking77-OOS \citep{banking77, oos_datasets} consists of $13,083$ customer service queries, categorized into $77$ fine-grained intents within the online banking domain. Among these, $50$ intents are in-scope, while the remaining $27$ are OOS queries. CLINC-OOS \citep{clinc, oos_datasets}, derived from the broader CLINC dataset, consists of two domains: "Banking" and "Credit cards", each featuring $10$ in-scope and $5$ OOS intents. The training sets for each domain include $500$ in-scope queries, while the test sets contain $850$ queries, with $350$ designated as OOS instances. Notably, our setting is unsupervised i.e., observations do not include intent labels for training. Further details are in App.~\ref{appendix:datasets}.

\noindent\textbf{Feature extractor \& class encoding.} We explored two feature encoders, namely the GTE model \citep{gte} and MPNET \citep{mpnet}, both pre-trained on a large corpus of text pairs across various domains. We found that directly encoding intent topic names using these encoders did not meet our performance expectations (See Sec.~\ref{sec:ablation_study}). To overcome this challenge, we leverage ChatGPT to generate a query corresponding to each topic and utilize these generated queries as class descriptions instead of the intent topic names. For further details, please refer to App.~\ref{appendix:zs}.

\noindent\textbf{Baselines.} We compare ColdFusion (Sec.~\ref{sec:coldfusion}) to several baselines. These include the zero-shot model (ZS), detailed in Sec.~\ref{sec:zs}, which relies solely on the generated normal class descriptions. Additionally, we consider DN2, an observation-based anomaly detection method proposed by \cite{panda}. DN2 computes the anomaly score of an observation by its deep $1$-nearest neighbor distance versus the previous observations $\mathcal{D}_{t}$. For implementation details refer to App.~\ref{appendix:implementation_details}.

\noindent\textbf{Evaluation metrics.} We propose a new metric to evaluate the cold-start setting, which emphasizes high-accuracy shortly after deployment (low $t$). At each time step $t$, we evaluate the performance of the anomaly scores model using the Area Under the Receiver Operation Characteristic (AUROC) curve. We obtain an AUROC score for every time step, and we denote them as $\{\text{AUC}(t)\}_{t=1}^{T}$. We summarize this $t$ vs. AUROC curve by the area under it up to time $t$. This is denoted as $\text{AUC}^2_{\tilde{t}}= \frac{\sum_{t'=1}^{t}\text{AUC}(t')}{t}$, where $\tilde{t} = \frac{t}{T}$, the fraction of the training set used. The $\text{AUC}^2_{\tilde{t}}$ metric provides a concise summary of the model's accuracy freshly after deployment.

\subsection{Results}
We present our results in Tab.~\ref{tab:co} and Fig.~\ref{fig:over_time}. ColdFusion consistently outperforms all baselines across the evaluated datasets by a large margin. Particularly, we see that DN2 performs poorly, especially with small $t$, and the zero-shot baseline (ZS) maintains constant performance over time. Conversely, our approach performs well even for low $t$ values, and improves over time. The presence of anomalies in the data stream poses a challenge for DN2, as it solely relies on the observed contaminated stream. This reliance often leads to occasional decreases in performance for DN2, highlighting the vulnerability of methods that exclusively depend on the observed data without considering the underlying anomalies. Furthermore, our method's robustness to different feature encoders, as evidenced by consistent trends in both the GTE and MPNET models, suggests that it is not reliant on a single feature extractor. Results for different contamination are in App.~\ref{appendix:results}.

\section{Ablation Study}
\label{sec:ablation_study}

\textbf{Class embedding adaptation method.} 
We investigate several variations of the adaptation method, shown in Tab.~\ref{tab:adaptation}. i) Replacing our assignment and adaptation stages with $K$-means notably reduces performance, mainly due to its less effective random initialization method vs. our descriptor initialization; ii) Iterating multiple steps of assignment and adaptation, each time assigning to the adapted center, fails to outperform ColdFusion. The single iteration of ColdFusion is preferred, since multiple iterations increase the computational cost. Additionally, the results in Tab.~\ref{tab:adaptation} show that median adaptation is slightly better than using the mean on the evaluated datasets.

\vspace{0.5em}

\noindent \textbf{Effectiveness of generated queries.} In Tab.~\ref{tab:zs}, we examine the impact of a naive ZS detector that simply encodes the intent names, compared to our ZS approach, which uses ChatGPT to generate a query for each intent and then encodes the generated query as the class embedding. The results highlight that naive encoding of intent names alone yields subpar performance, whereas our pre-processing procedure considerably improves results. 

\begin{table}[t]
\centering
\resizebox{1.0\linewidth}{!}{%
\begin{tabular}{llccc}
\toprule
 & {Method} & {B77} & {C-Bank} & {C-Cards} \\
 \cmidrule(lr){2-2} \cmidrule(lr){3-3} \cmidrule(lr){4-4} \cmidrule(lr){5-5}
% \midrule 
\parbox[t]{4mm}{\multirow{2}{*}{\rotatebox[origin=c]{90}{\scriptsize{GTE}}}} & {Naive} & {76.9} & {60.7} & {69.8} \\
& {Generated} & {\textbf{78.9}} & {\textbf{83.1}} & {\textbf{81.8}} \\
\cmidrule(lr){1-5}
\parbox[t]{4mm}{\multirow{2}{*}{\rotatebox[origin=c]{90}{\scriptsize{MPNET}}}} & {Naive} & {79.8} & {69.6} & {73.7} \\
& {Generated} & {\textbf{81.8}} & {\textbf{82.7}} & {\textbf{80.1}} \\
\bottomrule
\end{tabular}
}
\caption{Comparison of ZS models in terms of AUROC. As ZS models maintain constant performance over time and are not exposed to data, $\text{AUC}^2_{\tilde{t}}$ and contaminations are irrelevant. Best results are in bold.}
\label{tab:zs}
\end{table}

%% file: sec/5_conclusion.tex
\section{Conclusion}
We introduced the new setting of cold-start anomaly detection, modeling freshly deployed anomaly detection systems. Our proposed solution, ColdFusion, is a method for adapting zero-shot anomaly detection to align with an observation stream. We introduced an evaluation protocol and metrics for comparing future methods. 
\clearpage

%% file: sec/6_limitations.tex
\section*{Limitations}
Our proposed method has several limitations. i) Not all deployed anomaly detection systems encounter the generalized cold-start problem and indeed in the case where there are many observations and very few anomalies, it is sometimes better to use observation-driven methods e.g., DN2 \citep{panda}. However, we believe that it is a common issue, particularly in domains like chatbots; ii) Our approach relies on user-provided guidance for zero-shot detection, which is not always available; iii) We assume a low contamination ratio; if this ratio is significantly higher, the effectiveness of our method may decrease.

%% file: sec/7_ethics.tex
\section*{Ethics Statement}
Cold-start anomaly detection can have several benefits for society, including improving online services security by identifying fraudulent activities, unauthorized access and enhancing chatbot functionality by filtering out irrelevant queries early. However, it may also have some less positive use cases, such as being used for surveillance or profiling individuals without their consent. Our research is not geared towards these less positive use cases.

%% file: sec/8_ack.tex
\section*{Acknowledgements}
This research was partially supported by funding from IBM, the Israeli Science Foundation and the Israeli Council for Higher Education.

%% file: sec/X_appendix.tex
\section{Zero-Shot Anomaly Detection}
\label{appendix:zs}

Zero-shot (ZS) anomaly detection assigns an anomaly score $S(x)$ to each data point $x$ without relying on past data. Instead, it is guided by a set of class names $\{c_1, c_2,..., c_K\}$ provided by the user. To tackle this challenge, we leverage ChatGPT to generate a user query corresponding to each class topic name. We use these generated queries as class descriptions instead of the intent topic names.

\textbf{Query Generation:} Utilizing ChatGPT-3.5, we generate a user query for each topic to serve as our class descriptions. Here, [DOMAIN] represents the chatbot domain (e.g., "Banking"). We employ the following template: "Generate queries that someone would ask a chatbot in [DOMAIN]. Generate one-sentence queries for each of the following topics: $\{c_1, c_2, ..., c_K\}$." This process yields a set of $K$ user queries, denoted by $\{q_k\}_{k=1}^{K}$.

A pre-trained feature extractor $\phi$ maps each generated class name $q_k$ and observation $x$ to deep embeddings $\phi(q_k)$ and $\phi(x)$. Subsequently, we compute the $L_2$ distance between the example embeddings and each generated user query. The final anomaly score is determined by the distance to the nearest class:
\vspace{-0.25em}
\begin{equation*}
S_{zs}(x) = \min_k \{d(\phi(x),\phi(c_k))\}_{k=1}^K
\end{equation*}
Alg.~\ref{alg:zero_shot} outlines our zero-shot model.

A comparison between naive class names and generated queries is presented in Tab.~\ref{tab:zs}. 

\section{Experimental Details}
\label{sec:appendix}

\subsection{Datasets}
\label{appendix:datasets}
We employ three widely used datasets, Banking77-OOS and CLINC-OOS (which is split into CLINC-Banking and CLINC-Credit\_Cards), to evaluate our anomaly detection approach.

\textbf{Banking77-OOS.} Banking77-OOS \citep{banking77, oos_datasets} is an annotated intent classification dataset designed for online banking queries. Comprising $13,083$ customer service queries, each query is labeled with one of 77 fine-grained intents within the banking domain. The dataset focuses on fine-grained, single-domain intent detection. Of these 77 intents, Banking77-OOS incorporates 50 in-scope intents, while the out-of-scope (OOS) queries are constructed based on 27 held-out in-scope intents. The training set consists of 5,095 in-scope user queries, and the test set comprises 3,080 user queries, including 1,080 OOS instances. 

\textbf{CLINC-OOS.} CLINC-OOS \citep{clinc, oos_datasets} emanates from the broader CLINC dataset, encompassing 15 intent classes across 10 different domains, with integrated out-of-scope examples. For our evaluation, we focus on two domains: "Banking" and "Credit cards". Each domain is characterized by 5 in-scope and 10 out-of-scope intents. The training set for each domain comprises 500 in-scope user queries, while the test set includes 850 user queries, with 350 designated as out-of-scope instances.

\begin{algorithm}[t]
\caption{Zero-Shot Detector}
\label{alg:zero_shot}
\KwIn{$\mathcal{D}_{prior}$, $\phi$, query $x$.}
\KwOut{Anomaly score $S_{zs}(x)$.}
\textbf{Step 1:} Generate user queries using ChatGPT and $\mathcal{D}_{prior}$: $\{q_k\}_{k=1}^{K}$\;
\textbf{Step 2:} Encode generated queries: $\{\phi(q_k)\}_{k=1}^{K}$ and input query: $\phi(x)$\;
\textbf{Step 3:} Compute anomaly score for $x$: $S_{zs}(x) = \min_k \{d\phi(x),\phi(q_k))_{k=1}^K$\;
\end{algorithm}
\vspace{-0.25em}

\subsection{Implementation Details \& Baselines}
\label{appendix:implementation_details}
Our implementation relies on two feature encoders: the GTE model \citep{gte} and MPNET \citep{mpnet}, both pre-trained on a large corpus of text pairs across various domains. We use the \texttt{HuggingFace} library for both models. Specifically, for the GTE model, we employ the "thenlper/gte-large" model checkpoint, while for MPNET, we use the "sentence-transformers/all-mpnet-base-v2" model checkpoint. It's noteworthy that all baselines are using the same feature encoders in our comparisons. We use $L_2$ as a distance metric. For DN2 \citep{panda}, the implementation involves encoding $\mathcal{D}_t$ and the target query $x$ with our feature encoder $\phi$, followed by computing the $1$-nearest-neighbor distance to $\phi(\mathcal{D}_t)$. We employ the \texttt{faiss} library for nearest-neighbor distance computations. In our ColdFusion in order to be robust to anomalies, we excluded observations assigned to class $k$ but are further than $\tau$. Formally, we define $\mathcal{C}_k$, as the set of all observations assigned to class $k$ as:
\begin{equation*}
\label{eq:tau}
    \mathcal{C}_k = \{\phi(x)|a(x)=k, d(\phi(x),\phi(c_k)) \leq \tau\} 
\end{equation*}
We set $\tau$ by first computing the distances between all samples and their assigned centers, sorting them, and choosing $\tau$ as the $90\%$ percentile. An ablation study on this parameter is in App.~\ref{appendix:results}.

\begin{table*}[t]
\centering
\resizebox{1.0\linewidth}{!}{%
\begin{tabular}{llcccccccccccc}
\toprule
\parbox[t]{4mm}{\multirow{1}{*}{\rotatebox[origin=c]{90}{\small{Encoder}}}}  & \multirow{3}{*}{Method} & \multicolumn{3}{c}{$\text{AUC}^2_{10\%}$} & \multicolumn{3}{c}{$\text{AUC}^2_{25\%}$} & \multicolumn{3}{c}{$\text{AUC}^2_{50\%}$}& \multicolumn{3}{c}{$\text{AUC}^2_{100\%}$}  \\
  \cmidrule(lr){3-5} \cmidrule(lr){6-8} \cmidrule(lr){9-11} \cmidrule(lr){12-14}
  & & B77 & C-Bank & C-Cards &  B77 & C-Bank & C-Cards & B77 & C-Bank & C-Cards & B77 & C-Bank & C-Cards \\
 \cmidrule(lr){1-1} \cmidrule(lr){2-2} \cmidrule(lr){3-3} \cmidrule(lr){4-4} \cmidrule(lr){5-5} \cmidrule(lr){6-6} \cmidrule(lr){7-7} \cmidrule(lr){8-8} \cmidrule(lr){9-9} \cmidrule(lr){10-10} \cmidrule(lr){11-11} \cmidrule(lr){12-12} \cmidrule(lr){13-13} \cmidrule(lr){14-14}
% \midrule 
\parbox[t]{4mm}{\multirow{3}{*}{\rotatebox[origin=c]{90}{GTE}}} & ZS & \textbf{78.9} & 83.1 & 81.8 & 78.9 & 83.1 & 81.8 & 78.9 & 83.1 & 81.8 & 78.9 & 83.1 & 81.8  \\
& DN2 & 74.6 & 71.2 & 71.6 & 77.5 & 79.4 & 79.1 & 78.8 & 82.9 & 82.9 & 79.2 & 84.7 & 85.7 \\
& ColdFusion & 79.0 & \textbf{85.1} & \textbf{85.2}  & \textbf{80.9} & \textbf{86.9} & \textbf{87.6} & \textbf{81.8}  & \textbf{87.7} & \textbf{88.7} & \textbf{82.3} & \textbf{89.1} & \textbf{89.1} \\
\cmidrule(lr){1-14}
\parbox[t]{4mm}{\multirow{3}{*}{\rotatebox[origin=c]{90}{MPNET}}} & ZS & \textbf{81.8} & 82.7 & 80.1 & \textbf{81.8} & 82.7 & 80.1 & 81.8 & 82.7 & 80.1 & 81.8 & 82.7 & 80.1 \\
& DN2 & 76.6 & 74.7 & 70.7 & 79.1 & 82.4 & 78.5 & 80.1 & 85.7  & 82.7 & 80.5 & 86.9 & 84.8 \\
& ColdFusion & 80.6 & \textbf{87.0} & \textbf{85.2} & 81.7 & \textbf{89.0} & \textbf{87.6} & \textbf{82.5} & \textbf{89.5} & \textbf{89.0} & \textbf{83.2} & \textbf{90.0} & \textbf{89.1} \\
\bottomrule
\end{tabular}
}
% \end{center}
\caption{$\text{AUC}^2_{\tilde{t}}$ results, with contamination of $r=2.5\%$. Best results are in bold.}
\label{tab:co_2.5}
\vspace{-0.25em}
\end{table*}

\begin{table*}[t]
\centering
\resizebox{1.0\linewidth}{!}{%
\begin{tabular}{llcccccccccccc}
\toprule
\parbox[t]{4mm}{\multirow{1}{*}{\rotatebox[origin=c]{90}{\small{Encoder}}}}  & \multirow{3}{*}{Method} & \multicolumn{3}{c}{$\text{AUC}^2_{10\%}$} & \multicolumn{3}{c}{$\text{AUC}^2_{25\%}$} & \multicolumn{3}{c}{$\text{AUC}^2_{50\%}$}& \multicolumn{3}{c}{$\text{AUC}^2_{100\%}$}  \\
  \cmidrule(lr){3-5} \cmidrule(lr){6-8} \cmidrule(lr){9-11} \cmidrule(lr){12-14}
  & & B77 & C-Bank & C-Cards &  B77 & C-Bank & C-Cards & B77 & C-Bank & C-Cards & B77 & C-Bank & C-Cards \\
 \cmidrule(lr){1-1} \cmidrule(lr){2-2} \cmidrule(lr){3-3} \cmidrule(lr){4-4} \cmidrule(lr){5-5} \cmidrule(lr){6-6} \cmidrule(lr){7-7} \cmidrule(lr){8-8} \cmidrule(lr){9-9} \cmidrule(lr){10-10} \cmidrule(lr){11-11} \cmidrule(lr){12-12} \cmidrule(lr){13-13} \cmidrule(lr){14-14}
% \midrule 
\parbox[t]{4mm}{\multirow{3}{*}{\rotatebox[origin=c]{90}{GTE}}} & ZS & \textbf{78.9} & 83.1 & 81.8 & 78.9 & 83.1 & 81.8 & 78.9 & 83.1 & 81.8 & 78.9 & 83.1 & 81.8  \\
& DN2 & 70.6 & 67.2 & 71.3 & 72.5 & 77.5 & 78.1 & 73.4 & 80.5 & 81.1 & 73.8 & 80.8 &  81.9 \\
& ColdFusion & 77.4 & \textbf{83.4} & \textbf{86.4}  & \textbf{78.9} & \textbf{87.0} & \textbf{87.1} & \textbf{79.9}  & \textbf{88.4} & \textbf{87.9} & \textbf{80.8} & \textbf{88.9} & \textbf{88.2} \\
\cmidrule(lr){1-14}
\parbox[t]{4mm}{\multirow{3}{*}{\rotatebox[origin=c]{90}{MPNET}}} & ZS & \textbf{81.8} & 82.7 & 80.1 & \textbf{81.8} & 82.7 & 80.1 & 81.8 & 82.7 & 80.1 & 81.8 & 82.7 & 80.1 \\
& DN2 & 72.3 & 72.8 & 70.7 & 74.5 & 82.4 & 78.0 & 75.3 & 84.8  & 80.9 & 75.3 & 84.1 & 80.9 \\
& ColdFusion & 79.9 & \textbf{85.5} & \textbf{85.4} & 81.1 & \textbf{88.1} & \textbf{86.9} & \textbf{81.8} & \textbf{88.9} & \textbf{87.8} & \textbf{82.6} & \textbf{89.2} & \textbf{88.0} \\
\bottomrule
\end{tabular}
}
% \end{center}
\caption{$\text{AUC}^2_{\tilde{t}}$ results, with contamination of $r=7.5\%$. Best results are in bold.}
\label{tab:co_7.5}
\vspace{-0.25em}
\end{table*}

\section{More Results \& Analysis}
\label{appendix:results}

\begin{figure*}[ht]
     \centering
     \begin{subfigure}[b]{0.325\textwidth}
         \centering
         \includegraphics[width=\textwidth]{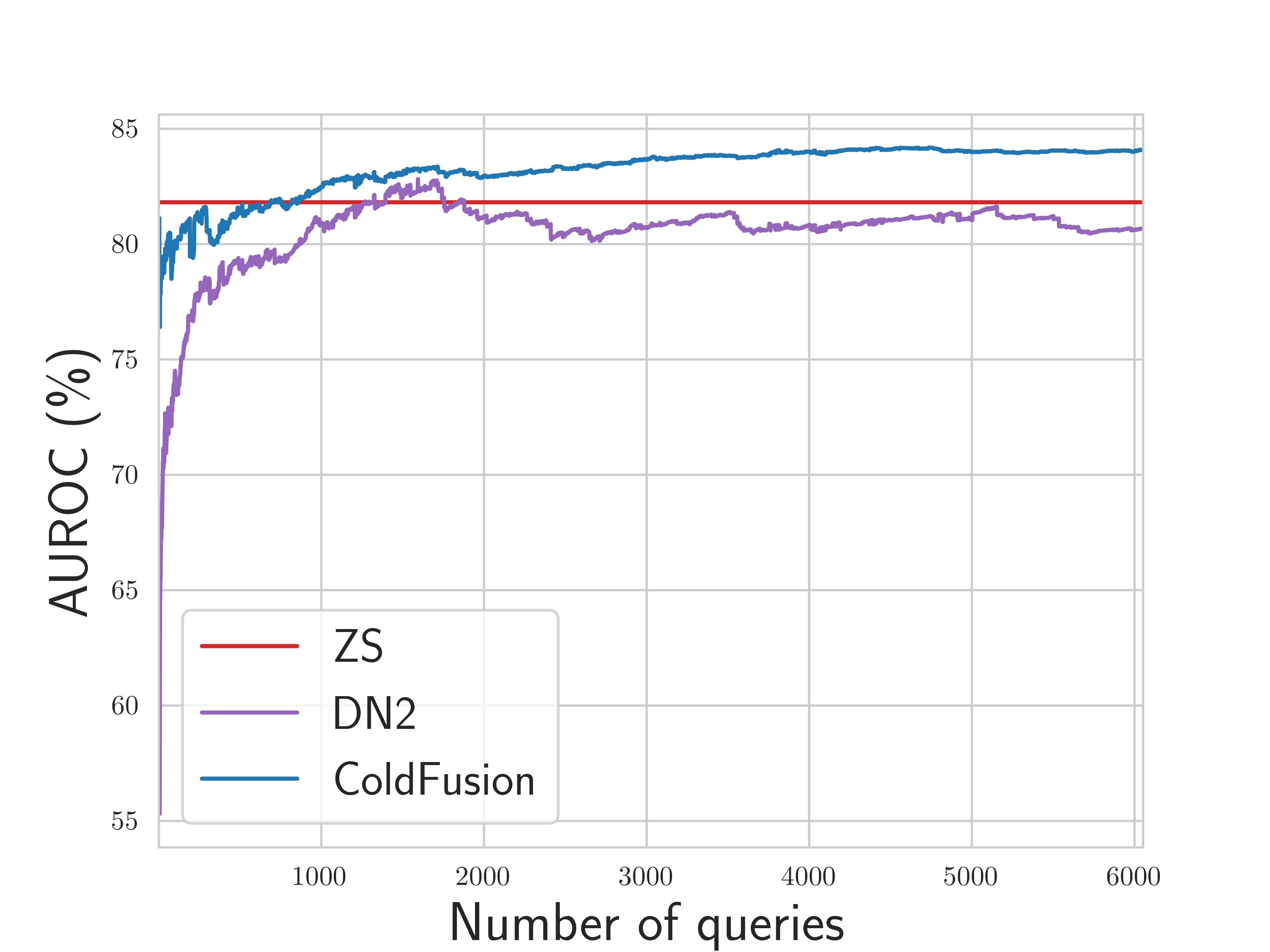}
         \vspace{-1em}
         \caption{Banking77-OOS}
     \end{subfigure}
     \hfill
     \begin{subfigure}[b]{0.325\textwidth}
         \centering
         \includegraphics[width=\textwidth]{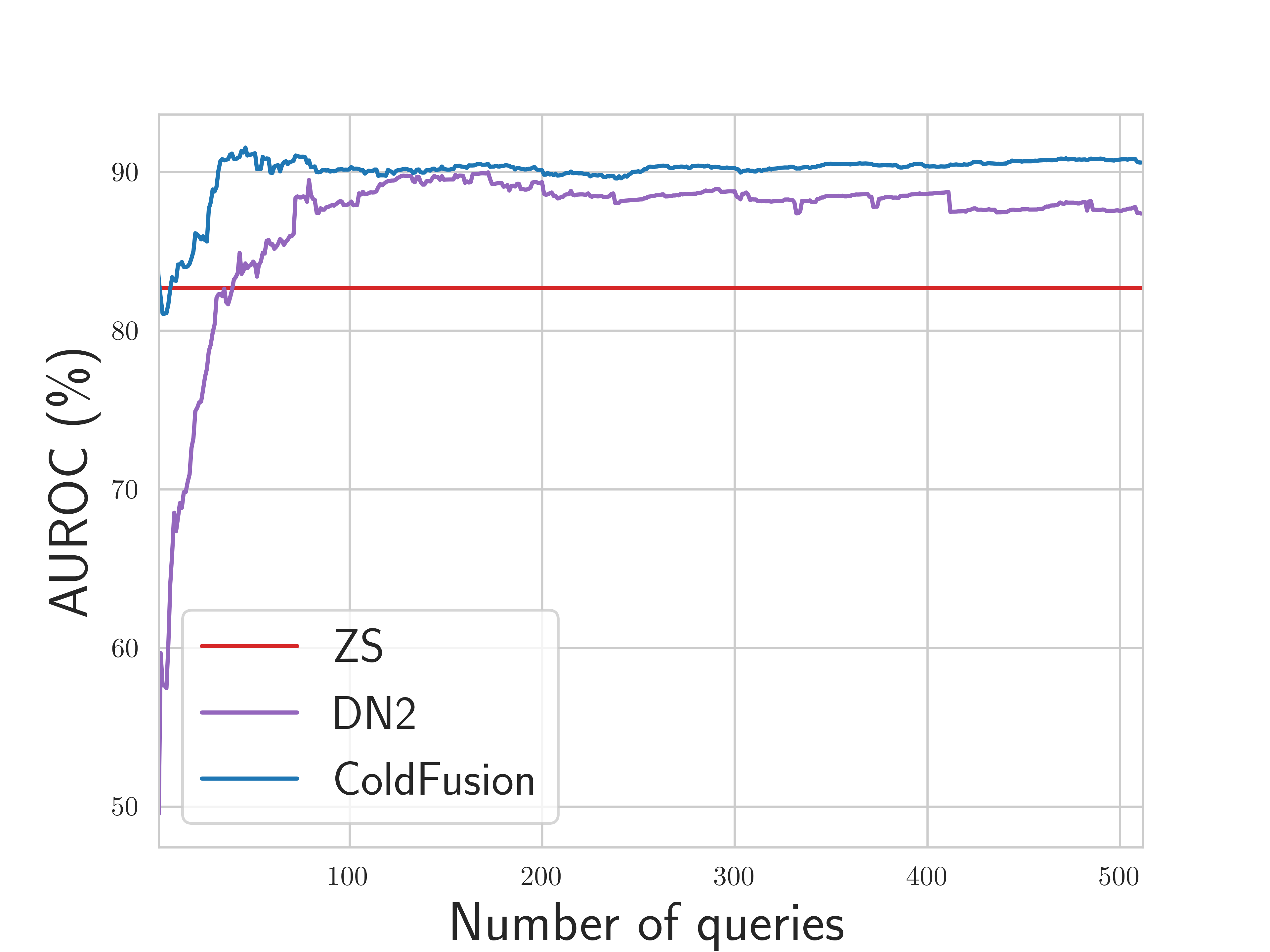}
         \vspace{-1em}
         \caption{CLINC-Banking}
     \end{subfigure}
     \hfill
     \begin{subfigure}[b]{0.325\textwidth}
         \centering
         \includegraphics[width=\textwidth]{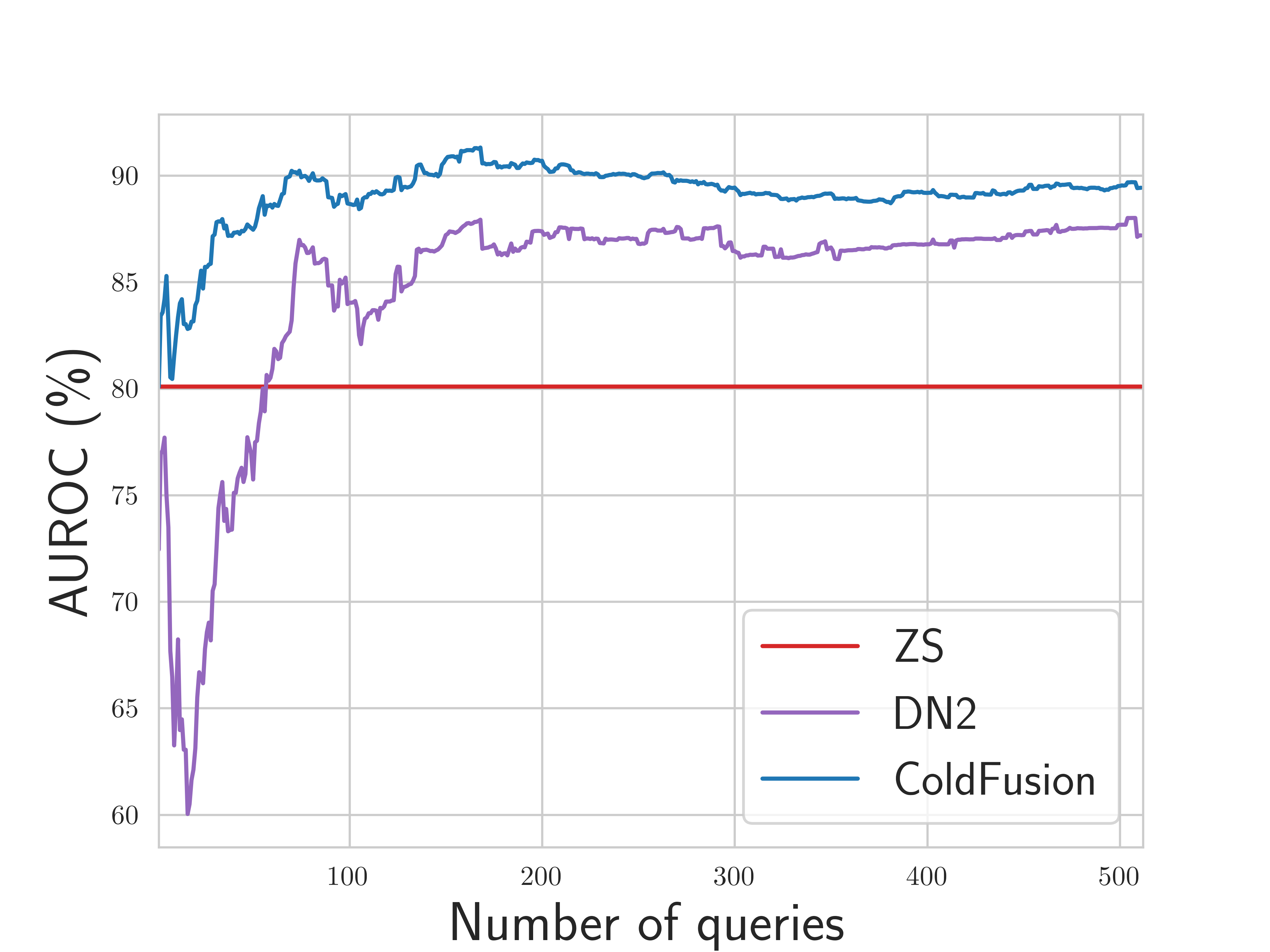}
         \vspace{-1em}
         \caption{CLINC-Credit\_Cards}
     \end{subfigure}
        \caption{Performance trends with contamination $r=2.5\%$ using the MPNET model over time.}
        \vspace{-0.25em}
        \label{fig:over_time_0.025_mpnet}
\end{figure*}

\begin{figure*}[ht]
     \centering
     \begin{subfigure}[b]{0.325\textwidth}
         \centering
         \includegraphics[width=\textwidth]{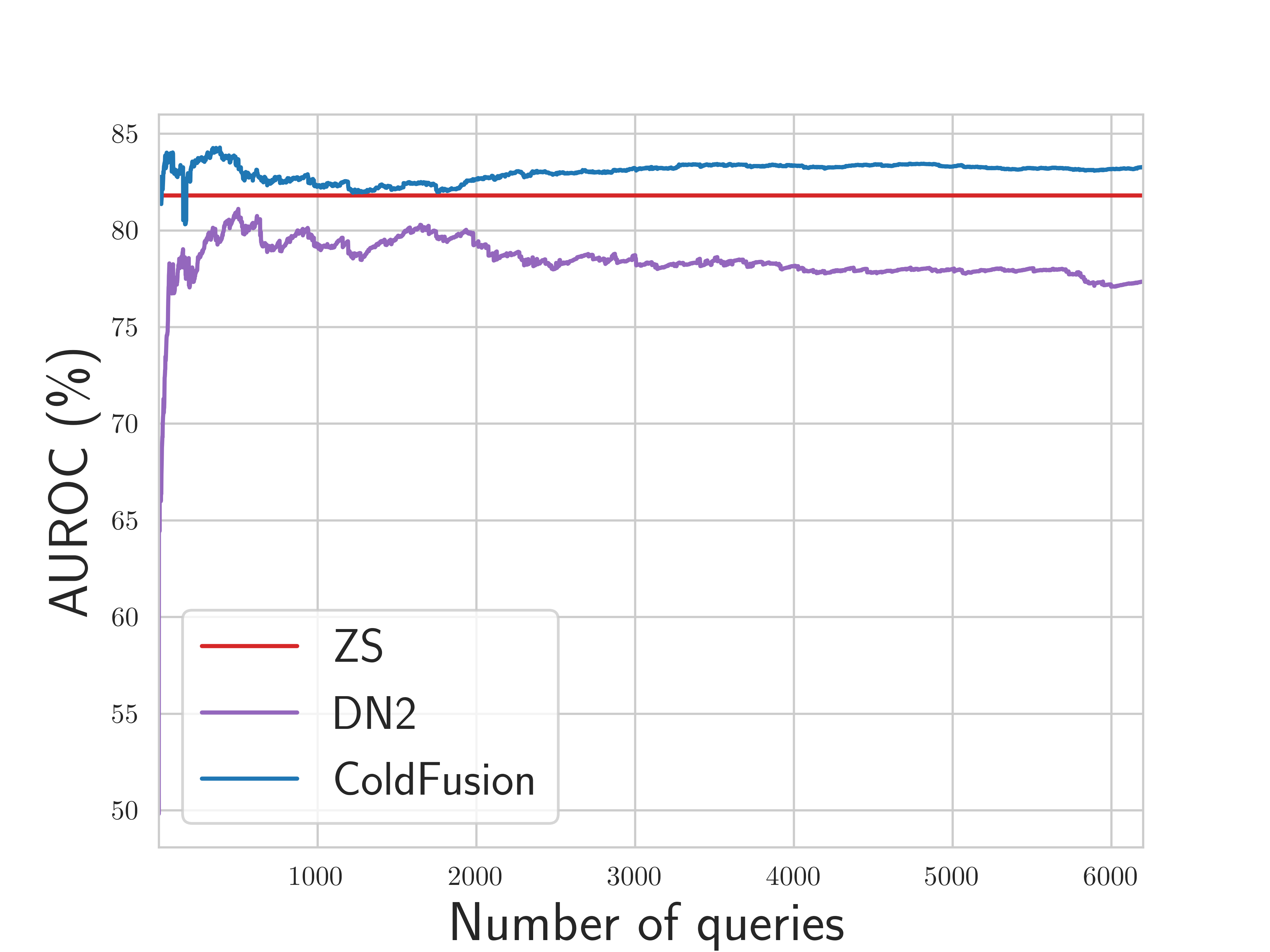}
         \vspace{-1em}
         \caption{Banking77-OOS}
     \end{subfigure}
     \hfill
     \begin{subfigure}[b]{0.325\textwidth}
         \centering
         \includegraphics[width=\textwidth]{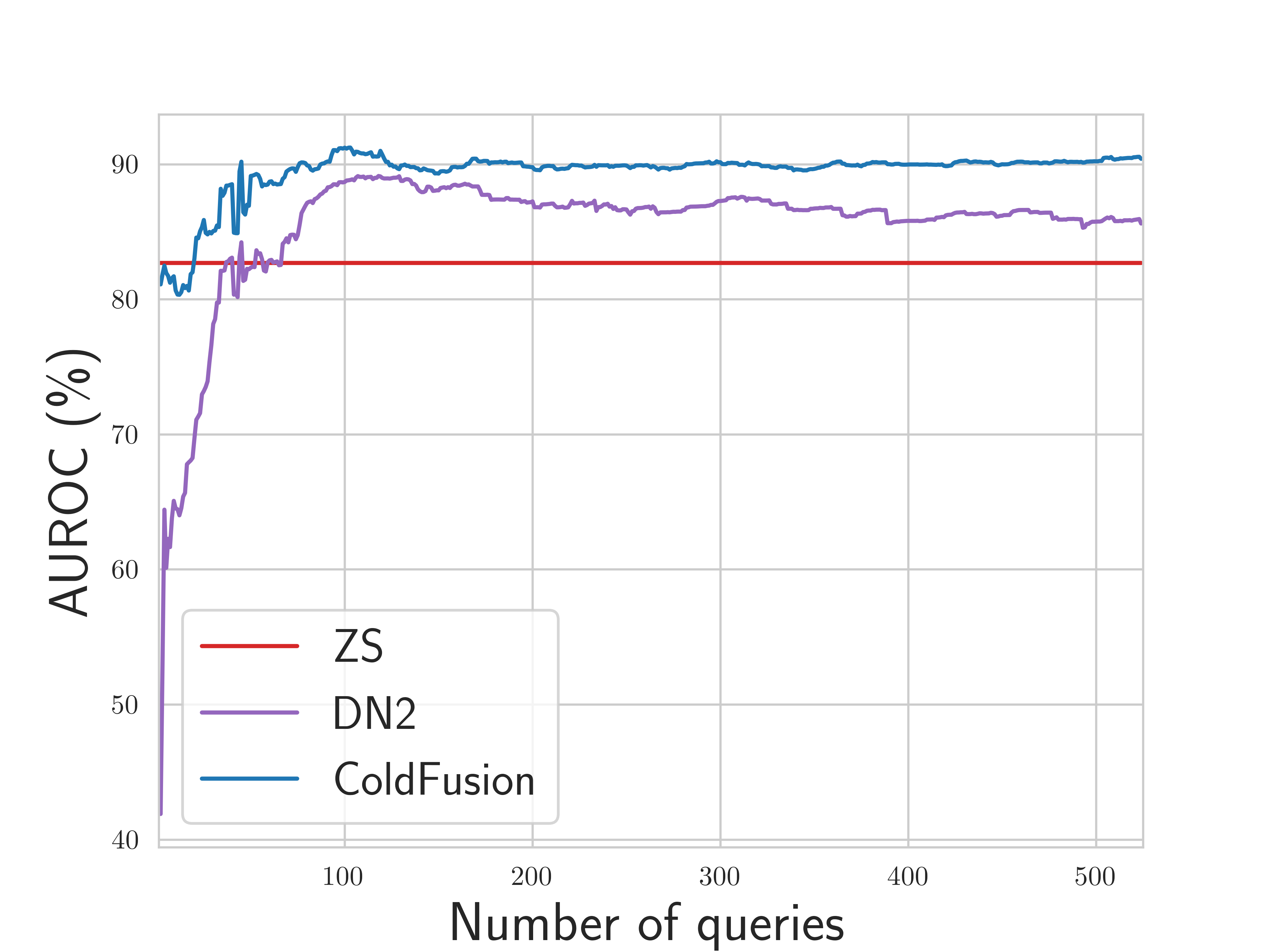}
         \vspace{-1em}
         \caption{CLINC-Banking}
     \end{subfigure}
     \hfill
     \begin{subfigure}[b]{0.325\textwidth}
         \centering
         \includegraphics[width=\textwidth]{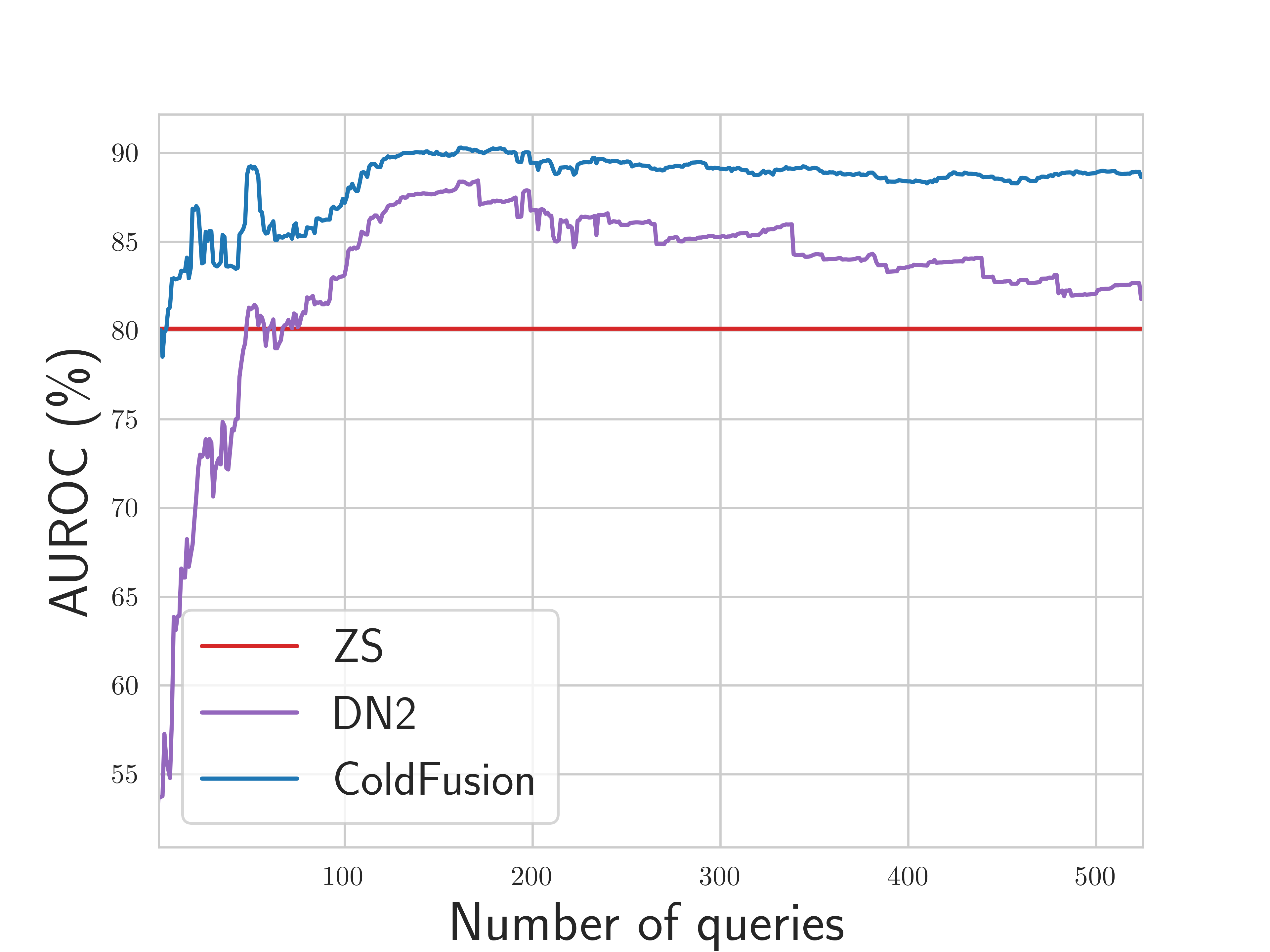}
         \vspace{-1em}
         \caption{CLINC-Credit\_Cards}
     \end{subfigure}
        \caption{Performance trends with contamination $r=5\%$ using the MPNET model over time.}
        \vspace{-0.25em}
        \label{fig:over_time_0.05_mpnet}
\end{figure*}

\begin{figure*}[ht]
     \centering
     \begin{subfigure}[b]{0.325\textwidth}
         \centering
         \includegraphics[width=\textwidth]{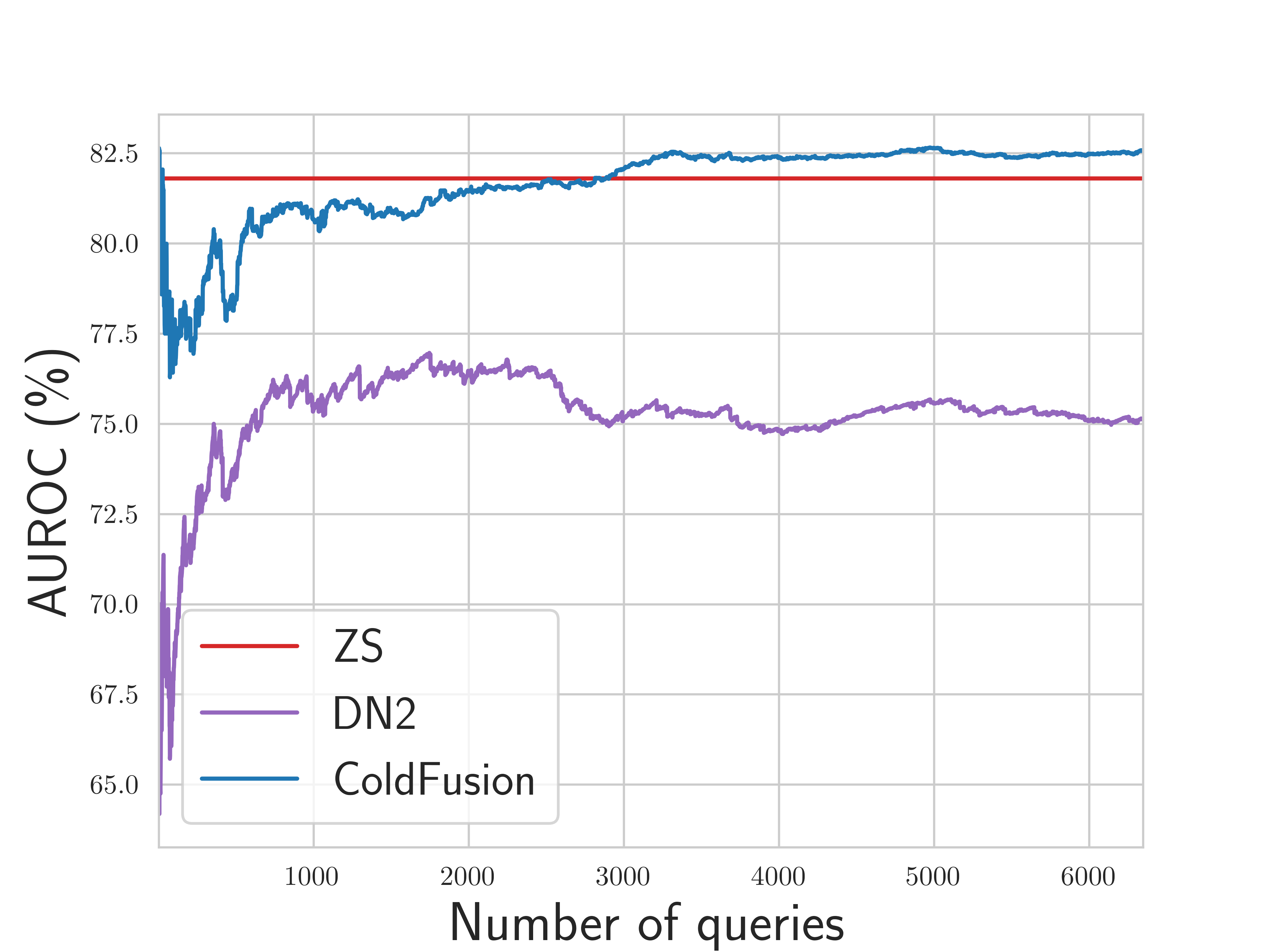}
         \vspace{-1em}
         \caption{Banking77-OOS}
     \end{subfigure}
     \hfill
     \begin{subfigure}[b]{0.325\textwidth}
         \centering
         \includegraphics[width=\textwidth]{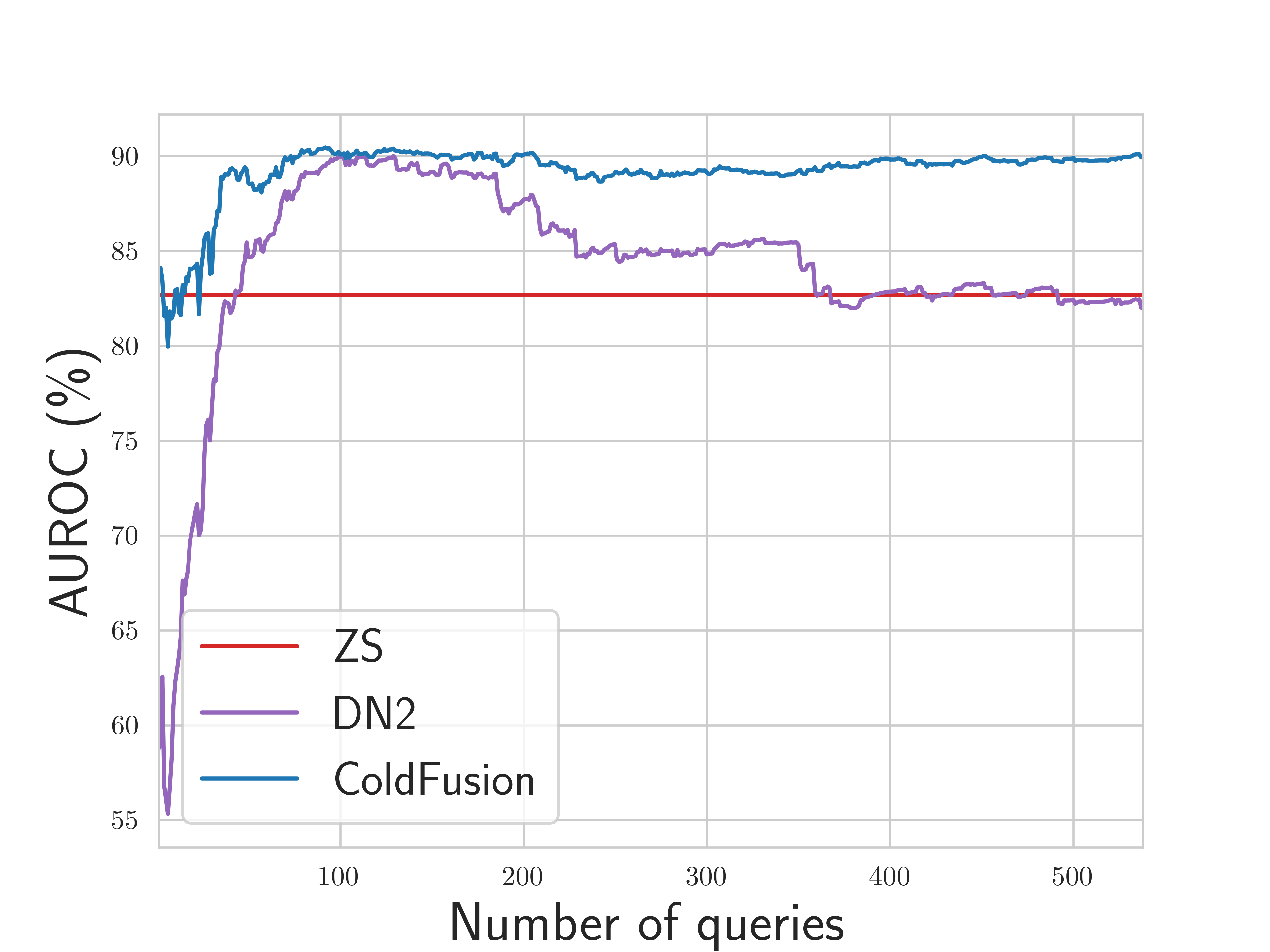}
         \vspace{-1em}
         \caption{CLINC-Banking}
     \end{subfigure}
     \hfill
     \begin{subfigure}[b]{0.325\textwidth}
         \centering
         \includegraphics[width=\textwidth]{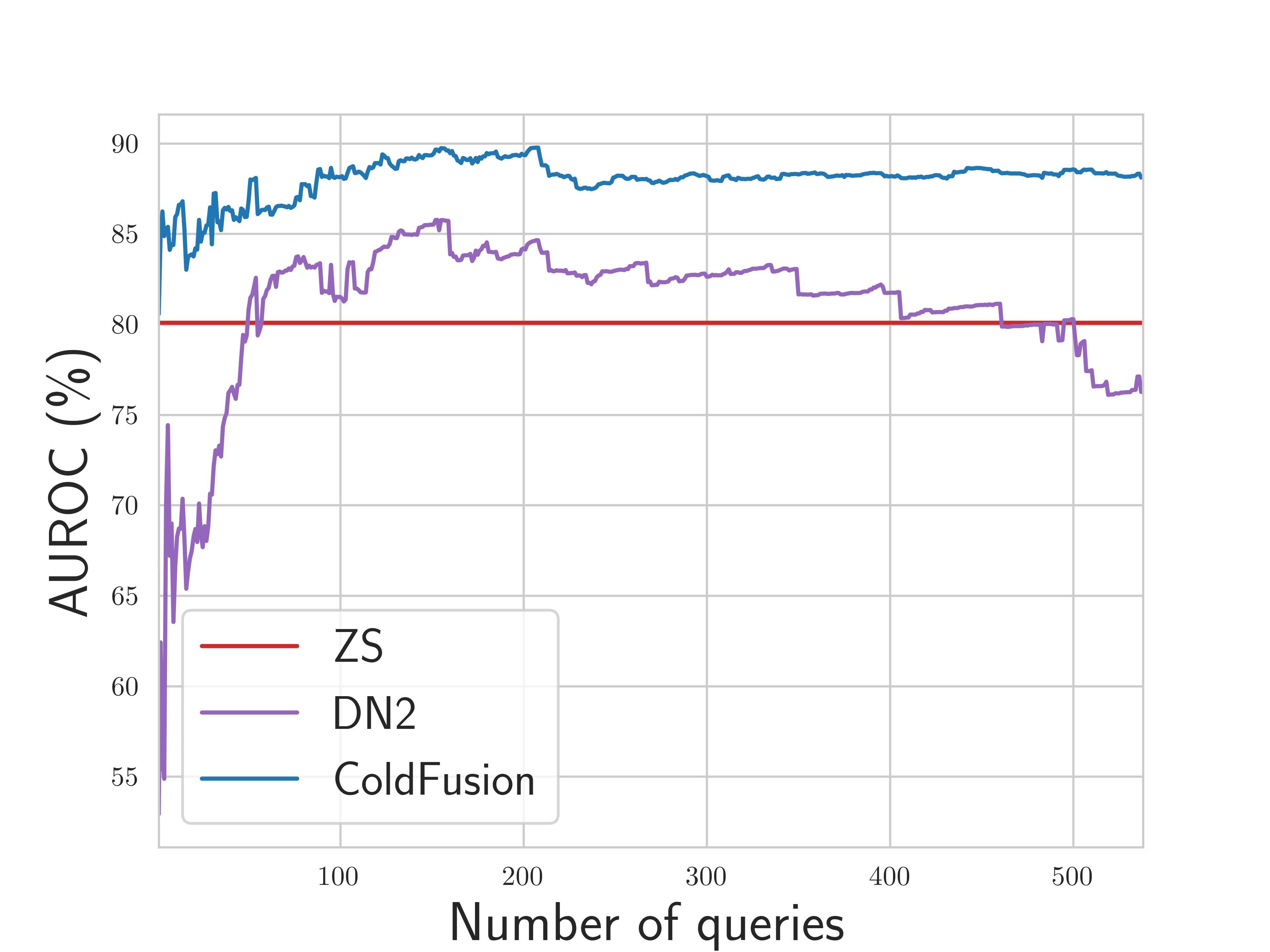}
         \vspace{-1em}
         \caption{CLINC-Credit\_Cards}
     \end{subfigure}
        \caption{Performance trends with contamination $r=7.5\%$ using the MPNET model over time.}
        \vspace{-0.25em}
        \label{fig:over_time_0.075_mpnet}
\end{figure*}

\begin{figure*}[ht]
     \centering
     \begin{subfigure}[b]{0.325\textwidth}
         \centering
         \includegraphics[width=\textwidth]{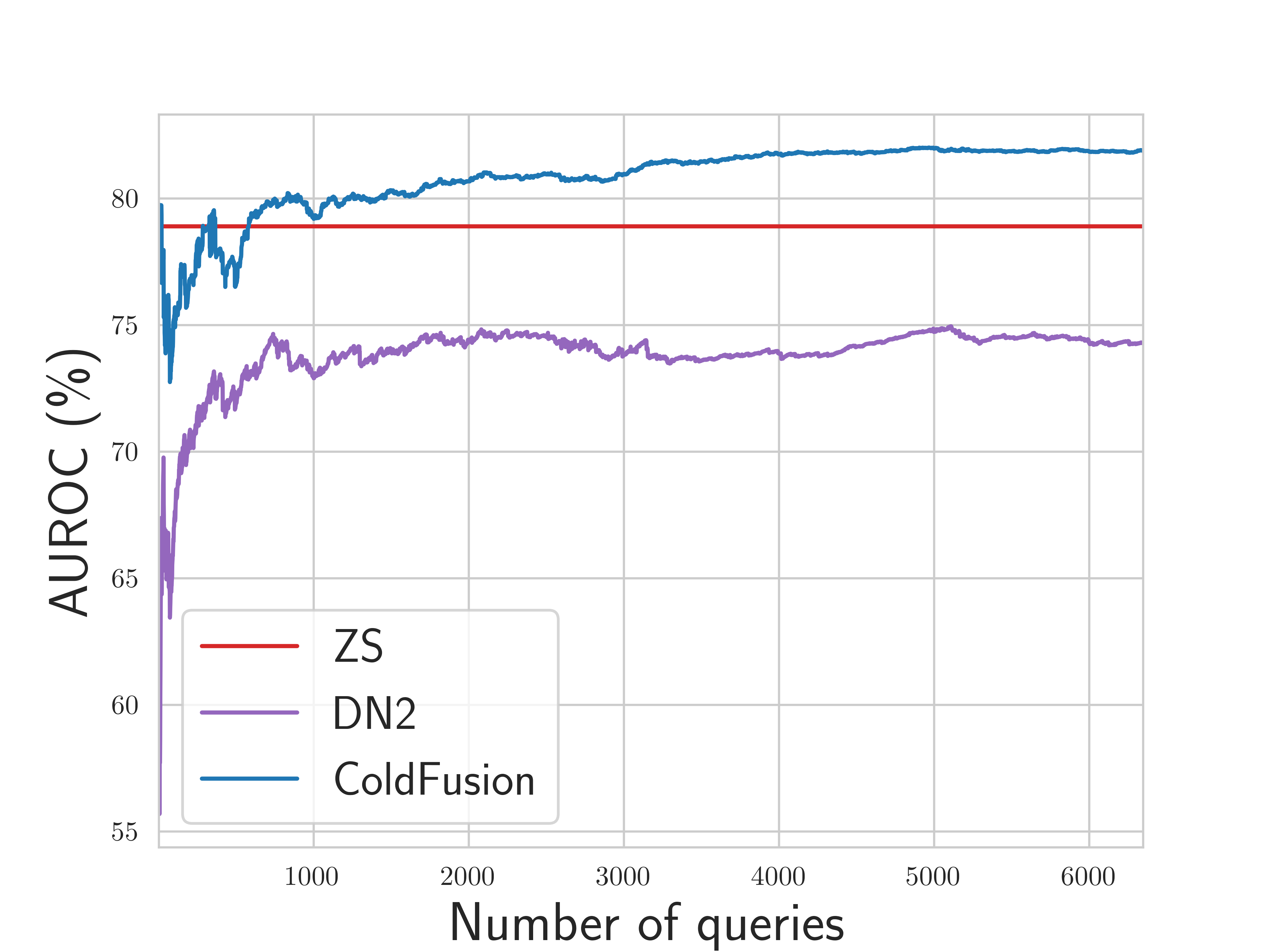}
         \vspace{-1em}
         \caption{Banking77-OOS}
     \end{subfigure}
     \hfill
     \begin{subfigure}[b]{0.325\textwidth}
         \centering
         \includegraphics[width=\textwidth]{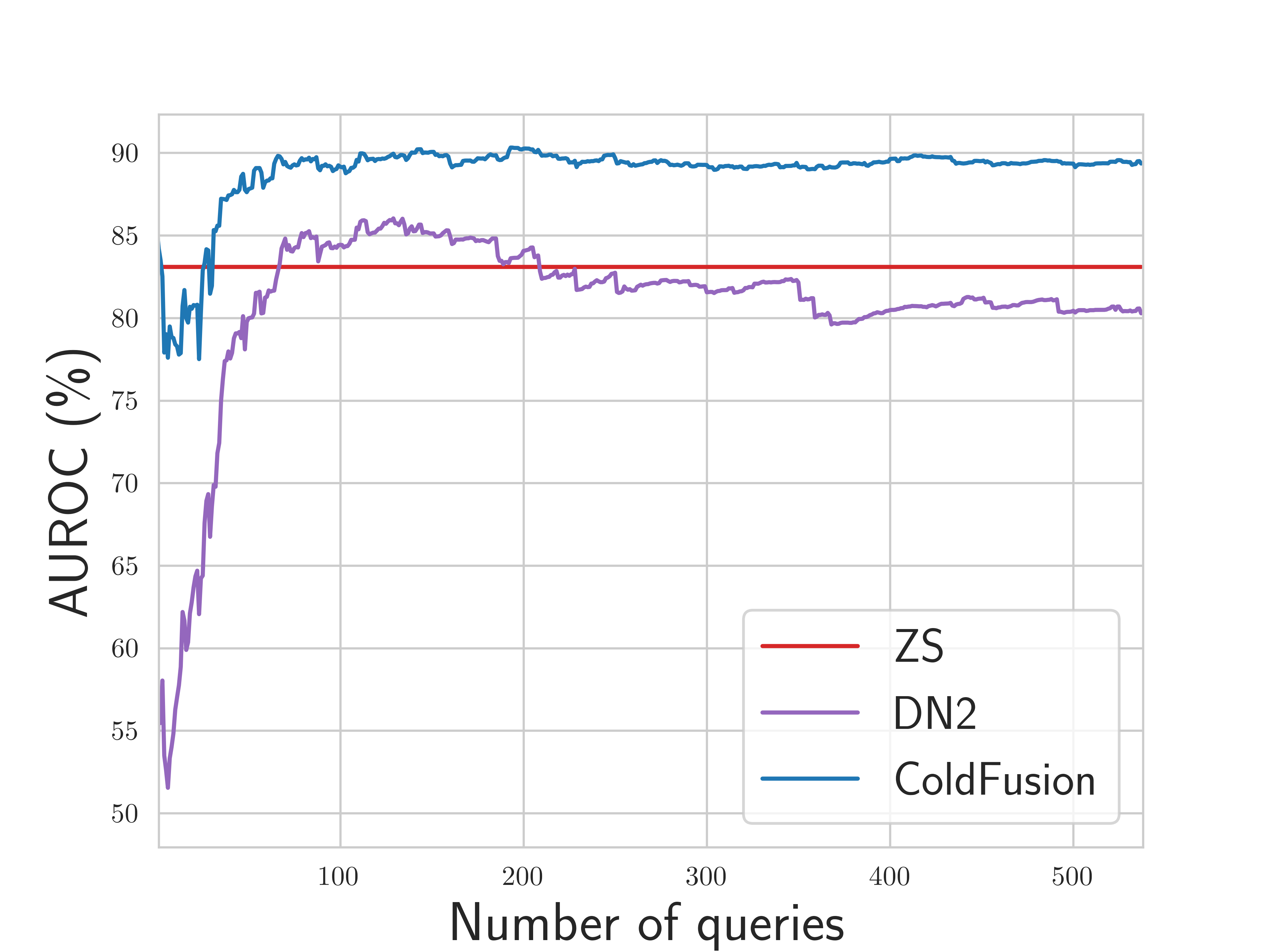}
         \vspace{-1em}
         \caption{CLINC-Banking}
     \end{subfigure}
     \hfill
     \begin{subfigure}[b]{0.325\textwidth}
         \centering
         \includegraphics[width=\textwidth]{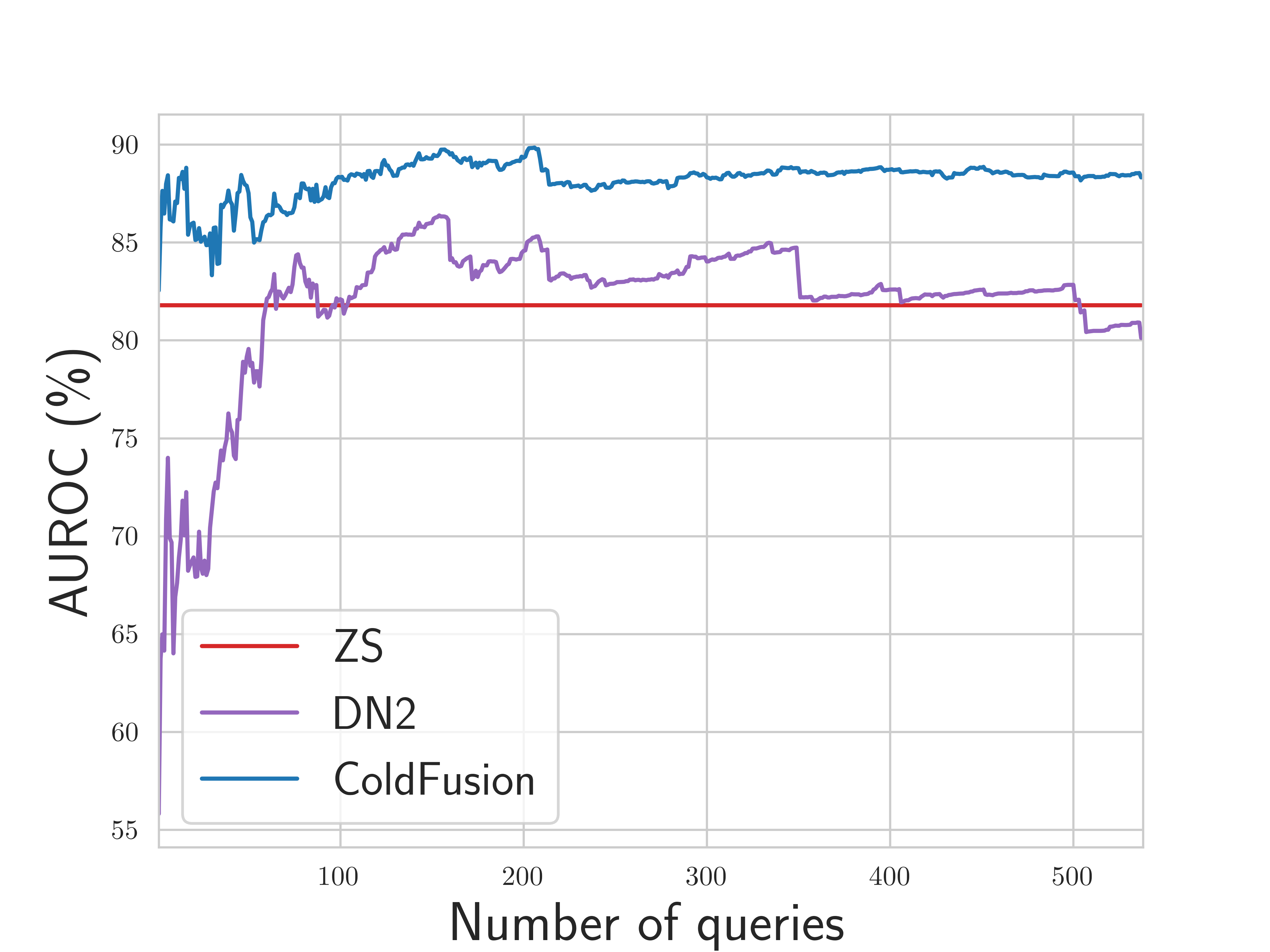}
         \vspace{-1em}
         \caption{CLINC-Credit\_Cards}
     \end{subfigure}
        \caption{Performance trends with contamination $r=7.5\%$ using the GTE model over time.}
        \vspace{-0.25em}
        \label{fig:over_time_0.075_gte}
\end{figure*}

\textbf{Contamination Ratios.} We extend our analysis by considering additional contamination ratios of $r\%=2.5$ and $r\%=7.5$, as shown in Tables~\ref{tab:co_2.5} and \ref{tab:co_7.5}, respectively. Additionally, we present visual insights into ColdFusion's adaptive performance over time through the figures presented in Fig.~\ref{fig:over_time_0.025_mpnet}, Fig.~\ref{fig:over_time_0.05_mpnet}, Fig.~\ref{fig:over_time_0.075_mpnet}, and Fig.~\ref{fig:over_time_0.075_gte}. Across all contamination ratios, ColdFusion consistently outperforms all baselines by a significant margin, reinforcing our approach's robustness and effectiveness. These supplementary results further support the stability and reliability of ColdFusion's performance trends observed in the main analysis.

\textbf{Effect of $\tau$.} Table~\ref{tab:tau} provides an ablation analysis of different $\tau$ parameters as defined in Eq.~\ref{eq:tau}. We observe that selecting the $50\%$ and $75\%$ percentiles yields suboptimal performance compared to using the $90\%$ and $100\%$ percentiles. These percentiles involve minimal filtering. Interestingly, there is a slight improvement in performance when employing the $90\%$ percentile compared to the $100\%$ percentile.

\section{Related Works}
\textbf{Out-of-scope intent discovery.} Out-of-scope (OOS) intent discovery involves clustering new, unknown intents to identify potential development directions and expand the capabilities of dialogue systems. Prior works \citep{lin2020discovering,zhang2021discovering,mou2022disentangled} in this domain have explored semi-supervised clustering using labeled in-domain data. Methods such as pre-training a BERT encoder with cross-entropy loss \citep{lin2020discovering,zhang2021discovering} and utilizing similarity constrained or supervised contrastive losses \citep{scl} to learn discriminative features \citep{mou2022disentangled} aim to transfer intent representations. However, these approaches face challenges related to in-domain overfitting, where representations learned from in-scope data may not generalize well to OOS data. In contrast to this line of work, our approach focuses on detecting OOS intents rather than discovering them. Notably, our setting involves unlabeled in-scope intents, and our model's prior knowledge is limited to intent names.

\textbf{Out-of-scope intent classification.} OOS intent classification is categorized based on the use of extensive labeled OOS intent samples during training. The first category involves methods that use OOS samples during training, treating OOS intent classification as a $(n+1)$-class classification task \citep{zheng2020out,zhan2021out}. In contrast, the second category aims to minimize intra-class variance and maximize inter-class variance to widen the margin between in-scope and OOS intents \citep{lin2019deep,zeng2021modeling}. Some approaches \citep{zhang2021deep,xu2020deep,zeng2021modeling} incorporate Gaussian distribution into the learned intent features to aid OOS detection. Our work stands apart from this line of research as it specifically addresses OOS intents, where in-scope intents (topics) lack labels, and the model has no information or exposure to any OOS intents.

\textbf{Classical anomaly detection methods.} Detecting anomalies in images has been researched for several decades. The methods follow three main paradigms: i) Reconstruction - this paradigm first attempts to characterize the normal data by a set of basis functions and then attempts to reconstruct a new example using these basis functions, typically under some constraint such as sparsity or weights with a small norm. Samples with high reconstruction errors are atypical of normal data distribution and anomalous. Some notable methods include: principal component analysis \citep{jolliffe2011principal} and $K$-nearest neighbors (kNN) \citep{eskin2002geometric}; ii) Density estimation - another paradigm is to first estimate the density of normal data. A new test sample is denoted as anomalous if its estimated density is low. Parametric density estimation methods include Ensembles of Gaussian Mixture Models (EGMM) \citep{glodek2013ensemble}, and non-parametric methods include $k$NN (which is also a reconstruction-based method) as well as kernel density estimation \citep{latecki2007outlier}. Both types of methods have weaknesses: parametric methods are sensitive to parametric assumptions about the nature of the data whereas non-parametric methods suffer from the difficulty of accurately estimating density in high-dimensions; iii) One-class classification (OCC) - this paradigm attempts to fit a parametric classifier to distinguish between normal training data and all other data. The classifier is then used to classify new samples as normal or anomalous. Such methods include one-class support vector machine (OCSVM) \citep{scholkopf2000support} and support vector data description (SVDD) \citep{tax2004support}.

\textbf{Deep learning for anomaly detection.} This line of work is based on the idea of initializing a neural network with pre-trained weights and then obtaining stronger performance by further adaptation of the training data. DeepSVDD \citep{ruff2018deep} suggested to first train an auto-encoder on the normal training data, and then using the encoder as the initial feature extractor. Moreover, since the encoder features are not specifically fitted to anomaly detection, DeepSVDD adapts to the encoder training data. However, this naive training procedure leads to catastrophic collapse. An alternative direction is to use features learned from auxiliary tasks on large-scale external datasets. Transferring pre-trained features for out-of-distribution detection has been proposed by \citep{hendrycks2019using}. It was recently established \citep{panda} that given sufficiently powerful representations, a simple criterion based on the kNN distance to the normal training data achieves strong performance. The best performing methods \citep{panda,reiss_eccvw,mean_shifted} combine pre-training on external datasets and a second finetuning stage on the provided normal samples in the training set, but they require many data observations and assume that the observations are not contaminated. To our knowledge, the only work that deals with a similar setting \citep{zero_winclip} assumes prior knowledge of anomalies, that observations come from a single normal class and that they are not contaminated by anomalies.

\begin{table*}[ht]
\centering
\resizebox{1.0\linewidth}{!}{%
\begin{tabular}{lcccccccccccc}
\toprule
 \multirow{3}{*}{\multirow{1}{*}{$\tau$}} & \multicolumn{3}{c}{$\text{AUC}^2_{10\%}$} & \multicolumn{3}{c}{$\text{AUC}^2_{25\%}$} & \multicolumn{3}{c}{$\text{AUC}^2_{50\%}$}& \multicolumn{3}{c}{$\text{AUC}^2_{100\%}$}  \\
  \cmidrule(lr){2-4} \cmidrule(lr){5-7} \cmidrule(lr){8-10} \cmidrule(lr){11-13}
  & B77 & C-Bank & C-Cards &  B77 & C-Bank & C-Cards & B77 & C-Bank & C-Cards & B77 & C-Bank & C-Cards \\
 \cmidrule(lr){1-1} \cmidrule(lr){2-2} \cmidrule(lr){3-3} \cmidrule(lr){4-4} \cmidrule(lr){5-5} \cmidrule(lr){6-6} \cmidrule(lr){7-7} \cmidrule(lr){8-8} \cmidrule(lr){9-9} \cmidrule(lr){10-10} \cmidrule(lr){11-11} \cmidrule(lr){12-12} \cmidrule(lr){13-13}
$\tau=perc(\phi(\mathcal{D}_t), 50\%)$ & 80.1 & 80.4 & 80.7 & 80.6 & 83.6 & 83.0 & 80.7 & 85.1 & 84.6 & 81.3 & 86.5 & 85.4  \\
$\tau=perc(\phi(\mathcal{D}_t), 75\%)$ & 81.8 & 81.9 & 83.0 & 81.9 & 85.5 & 84.8 & 81.8 & 87.4 & 86.7 & 82.1 & 88.2 & 87.7  \\
$\tau=perc(\phi(\mathcal{D}_t), 100\%)$ & \textbf{82.0} & 81.1 & \textbf{85.0} & \textbf{82.1} & 86.0 & 86.7 & 81.8 & 88.0 & 88.0 & \textbf{82.3} & 89.0 & 88.5  \\
$\tau=perc(\phi(\mathcal{D}_t), 90\%)$ & 81.7 & \textbf{82.3} & 84.8 & 81.8 & \textbf{87.0} & \textbf{87.3} & \textbf{81.9} & \textbf{88.6} & \textbf{88.7} & \textbf{82.3} & \textbf{89.2} & \textbf{89.0}  \\
\bottomrule
\end{tabular}
}
\caption{$\text{AUC}^2_{\tilde{t}}$ results using the GTE model, with contamination of $r=5\%$. Best results are in bold.}
\label{tab:tau}
\end{table*}